%% file: main.tex
\documentclass[10pt,twocolumn,letterpaper]{article}

\usepackage[pagenumbers]{cvpr} 
\usepackage{makecell}
\usepackage{multirow}
\usepackage{graphicx}

\usepackage[table]{xcolor}
\usepackage{xcolor}
\usepackage{adjustbox}
\usepackage{appendix}


\definecolor{cvprblue}{rgb}{0.21,0.49,0.74}
\usepackage[pagebackref,breaklinks,colorlinks,allcolors=cvprblue]{hyperref}

\title{Semantic Equitable Clustering: A Simple and Effective Strategy for Clustering Vision Tokens}

\author{%
  Qihang Fan $^{1, 2}$, Huaibo Huang$^{1}$\thanks{Huaibo Huang is the corresponding author.}, Mingrui Chen$^{1, 2}$, Ran He$^{1, 2}$\\
  $^1$MAIS \& NLPR, Institute of Automation, Chinese Academy of Sciences, Beijing, China\\
  $^2$School of Artificial Intelligence, University of Chinese Academy of Sciences, Beijing, China\\
  \texttt{fanqihang.159@gmail.com, huaibo.huang@cripac.ia.ac.cn,}\\
  \texttt{charmier@hust.edu.cn, rhe@nlpr.ia.ac.cn}\\
}
\begin{document}
\maketitle
\input{sec/0_abstract}    
\input{sec/1_intro}
\input{sec/2_related}
\input{sec/3_method}
\input{sec/4_experiments}

\input{sec/5_conclusion}
\input{sec/6_ack}
\input{sec/X_suppl}

\clearpage

{
    \small
    \bibliographystyle{ieeenat_fullname}
    \bibliography{main}
}


\end{document}

%% file: sec/0_abstract.tex
\begin{abstract}
The Vision Transformer (ViT) has gained prominence for its superior relational modeling prowess. However, its global attention mechanism's quadratic complexity poses substantial computational burdens. A common remedy spatially groups tokens for self-attention, reducing computational requirements. Nonetheless, this strategy neglects semantic information in tokens, possibly scattering semantically-linked tokens across distinct groups, thus compromising the efficacy of self-attention intended for modeling inter-token dependencies. Motivated by these insights,  we introduce a fast and balanced clustering method, named 
 \textbf{S}emantic \textbf{E}quitable \textbf{C}lustering (SEC).  SEC clusters tokens based on their global semantic relevance in an efficient, straightforward manner.  In contrast to traditional clustering methods requiring multiple iterations, our method achieves token clustering in a single pass. Additionally, SEC regulates the number of tokens per cluster, ensuring a balanced distribution for effective parallel processing on  current computational platforms without necessitating further optimization. Capitalizing on SEC, we propose a versatile vision backbone, SECViT. Comprehensive experiments in image classification, object detection, instance segmentation, and semantic segmentation validate the effectiveness of SECViT. Moreover, SEC can be conveniently and swiftly applied to multimodal large language models (MLLM), such as LLaVA, to serve as a vision language connector, effectively accelerating the model's efficiency while maintaining unchanged or better performance.
\end{abstract}

%% file: sec/1_intro.tex
\section{Introduction}
\label{sec:intro}

Since its inception, the Vision Transformer (ViT)\citep{vit} has drawn considerable interest from the research community due to its robust modeling prowess. However, the quadratic complexity of Self-Attention leads to significant computational overhead, thus constraining the practicality of ViT. A variety of strategies have been devised to alleviate this computational load, the most prevalent of which involves token grouping, thereby constraining the attention span of each token\citep{SwinTransformer, cswin, crossformer, maxvit}.


Specifically, the Swin-Transformer~\citep{SwinTransformer} partitions tokens into multiple small windows, restricting token attention within each window. The CSWin-Transformer~\citep{cswin} adopts a cross-shaped grouping, endowing each token with a global receptive field. MaxViT~\citep{maxvit} amalgamates window and grid attention, facilitating intra-window tokens to attend to their counterparts in other windows.
However, these methods, solely reliant on spatial positioning, neglect token semantics, potentially restricting the self-attention's capacity to model semantic dependencies.
To mitigate this, DGT~\citep{DGT} employs k-means clustering for query grouping, considering the semantic information of tokens for enhanced feature learning.
Nonetheless, the iterative nature of k-means clustering and the potential for uneven token counts per cluster can impact the efficiency of parallel attention operations.

\begin{figure}[t]
    \centering
    \includegraphics[width=0.99\linewidth]{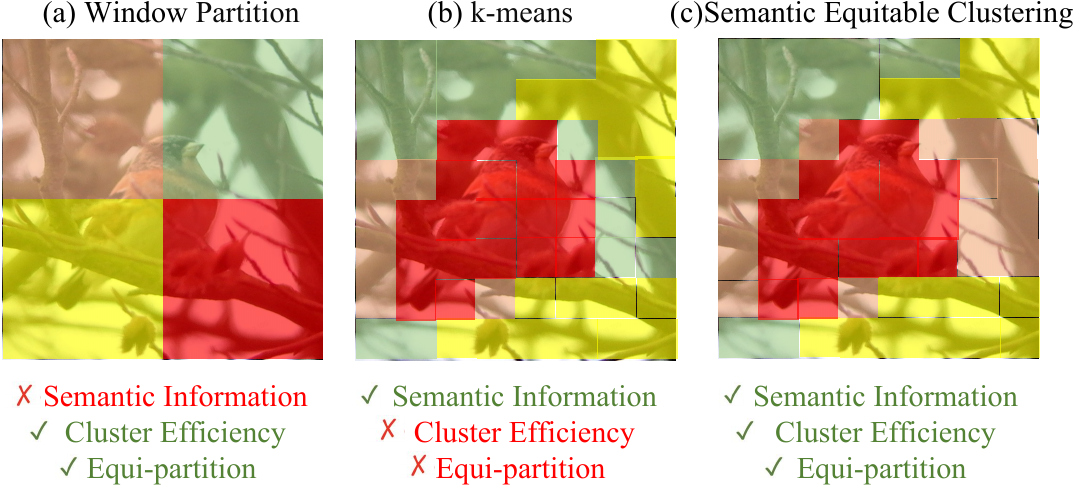}
    \vspace{-3mm}
    \caption{Comparison among Window Partition, Dynamic Group by k-means, and 
 Semantic Equitable Clustering. Our Semantic Equitable Clustering incorporates image semantics while maintaining efficient clustering, eliminating the need for iterative processes such as in k-means. Furthermore, it enables equi-partitioning of tokens, promoting efficient GPU processing without necessitating additional CUDA optimization.}
    \vspace{-3mm}
    \label{fig:intro}
\end{figure}

Given these considerations, an optimal token partitioning scheme should efficiently segregate tokens, incorporate semantic information, and efficiently utilize computational resources (e.g., GPU). In response, we introduce a simple, fast and equitable clustering approach named Semantic Equitable Clustering (SEC). SEC segments tokens based on their relevance to global semantic information. Specifically, we employ global pooling to generate a global token encapsulating global semantic information. The similarity between this global token and all other tokens is then computed, reflecting global semantic relevance. Upon obtaining the similarity matrix, tokens (excluding the global token) are sorted by similarity scores, and the tokens with similar scores are grouped into clusters, ensuring uniform token distribution across clusters. As depicted in Fig.~\ref{fig:intro}, SEC comprehensively considers token semantics and completes the clustering process in a single iteration, unlike the multi-iteration k-means. The resulting clusters, containing an equal number of tokens, can be processed in parallel by the GPU efficiently. 

\begin{figure}[t]
    \centering
    \includegraphics[width=0.99\linewidth]{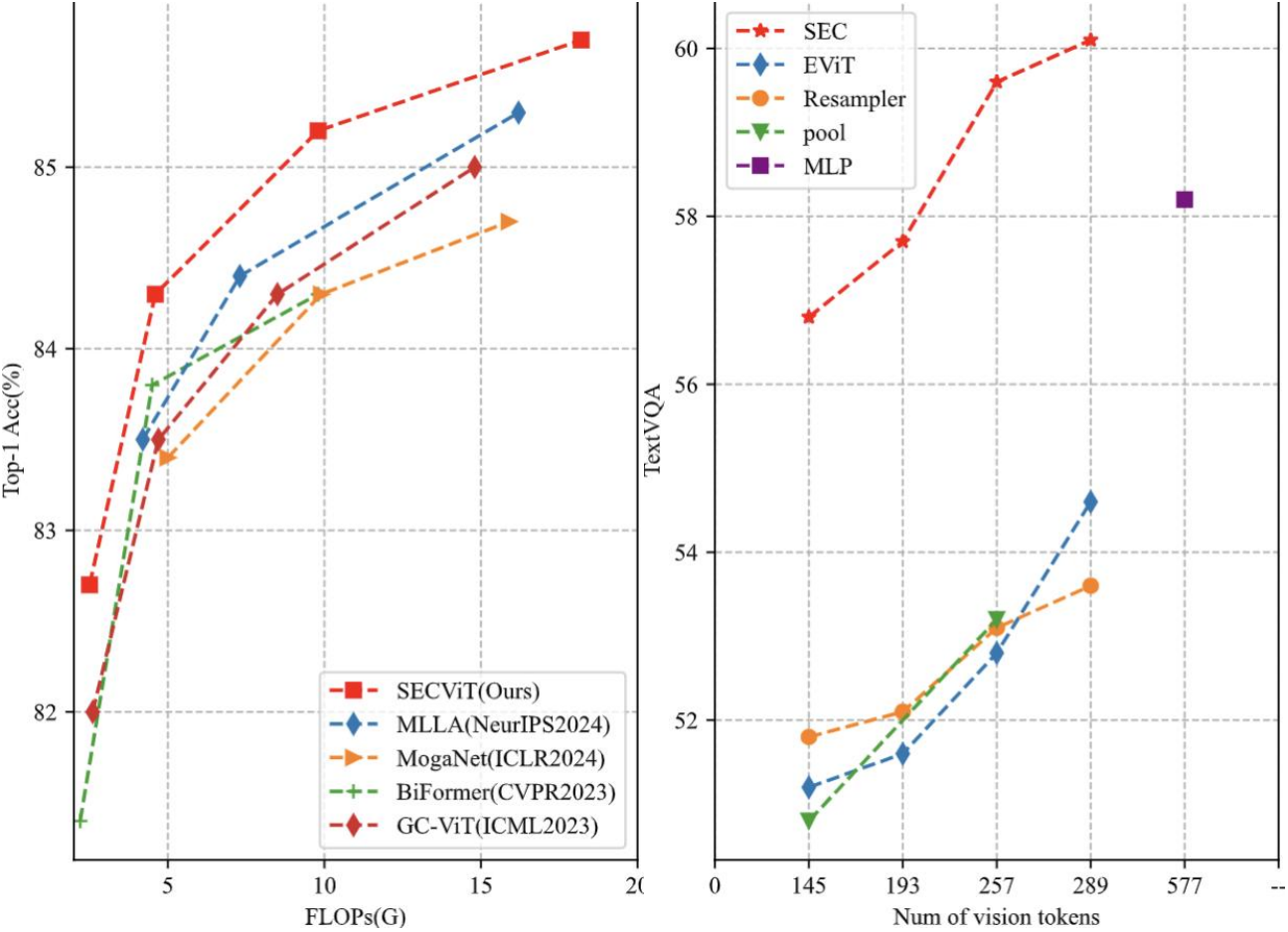}
    \vspace{-3mm}
    \caption{\textbf{Left: }Top-1 accuracy v.s. FLOPs on ImageNet-1K of recent SOTA models. \textbf{Right: }Comparison among different vision language connectors on LLaVA-1.5}
    \label{fig:comp}
    \vspace{-4mm}
\end{figure}


Building upon Semantic Equitable Clustering (SEC), we introduce the Semantic Equitable Clustering Vision Transformer (SECViT), a versatile vision backbone that is adaptable to a wide spectrum of downstream tasks. As demonstrated in Fig.~\ref{fig:comp}, SECViT exhibits significant performance improvements compared to previous state-of-the-art (SOTA) models. Impressively, SECViT attains an accuracy of \textbf{84.3\%} utilizing merely \textbf{4.6G}FLOPS, without the need for additional training data or supervision. This superior performance is maintained across different model scales. Furthermore, SECViT proves its proficiency in downstream tasks, including but not limited to, object detection, instance segmentation, and semantic segmentation. 

Beyond vision tasks, we also apply SEC to multimodal large language models (MLLM) such as LLaVA-1.5~\citep{llava1.5} to serve as an efficient vision language connector. Specifically, we use SEC to cluster the vision tokens, and then merge all the tokens at corresponding positions within each cluster into a single token. Experiments demonstrate that this approach significantly enhances the efficiency of LLaVA-1.5 while improving the model's performance.


%% file: sec/2_related.tex
\section{Related Works}
\label{sec:related_work}
\paragraph{Vision Transformer.}The Vision Transformer (ViT)~\citep{vit} is considered a powerful visual architecture. Many works have improved the Vision Transformer, including enhancing its training efficiency and reducing its computational cost~\citep{SwinTransformer, cswin, deit, tokenlabel, biformer}. DeiT~\citep{deit} uses distillation loss and incorporates extensive data augmentation methods into the ViT training process. Hierarchical structures represented by PVT~\citep{pvt, pvtv2, iformer, cmt, dat} reduce the number of tokens in global attention by downsampling the keys and values (KV), thereby low the computational cost. In addition to them, some methods directly prune tokens based on their importance, retaining important tokens~\citep{dynamicvit, evit}. This reduces the number of tokens and subsequently lowers the computational cost of the model. Another highly representative approach is to group all tokens such that each token can only attend to tokens within its own group~\citep{SwinTransformer, cswin, biformer, davit, DGT}. This method also significantly reduces the computational cost of self-attention.
\vspace{-3mm}

\paragraph{Grouping-Based Vision Transformer.}Most grouping-based attention mechanisms perform grouping based on spatial structure~\citep{SwinTransformer, cswin, maxvit, davit, DGT}. Specifically, the Swin-Transformer~\citep{SwinTransformer} divides all tokens into equally sized windows based on their spatial positions, where each token can only attend to tokens within its own window. This significantly reduces the model's computational cost.  In addition to dividing tokens into small windows along the spatial dimension, DaViT~\citep{davit} also splits channels into multiple groups along the channel dimension. Unlike the above methods that only consider positional information for grouping, DGT~\citep{DGT} takes semantic information into account by using k-means clustering to group the queries.
\vspace{-3mm}

\paragraph{Vision Language Connector.}The vision language connector is a critical component in MLLMs~\citep{llava1.5, honeybee, perceiver}. It aligns vision tokens with language tokens. Typical vision language connectors include MLP~\citep{llava1.5}, Resampler~\citep{bai2023qwenvl}, C-Abstractor~\citep{honeybee}, and others. Although MLP performs well, it introduces a significant number of vision tokens, which hampers the model's efficiency. On the other hand, connectors like Resampler improve the model's efficiency, but at the cost of reduced performance. Unlike these methods, our proposed SEC consider the semantic information of each token and significantly enhances the model's efficiency while maintaining its performance. 
\vspace{-3mm}

\begin{figure*}[!ht]
    \centering
    \includegraphics[width=0.95\linewidth]{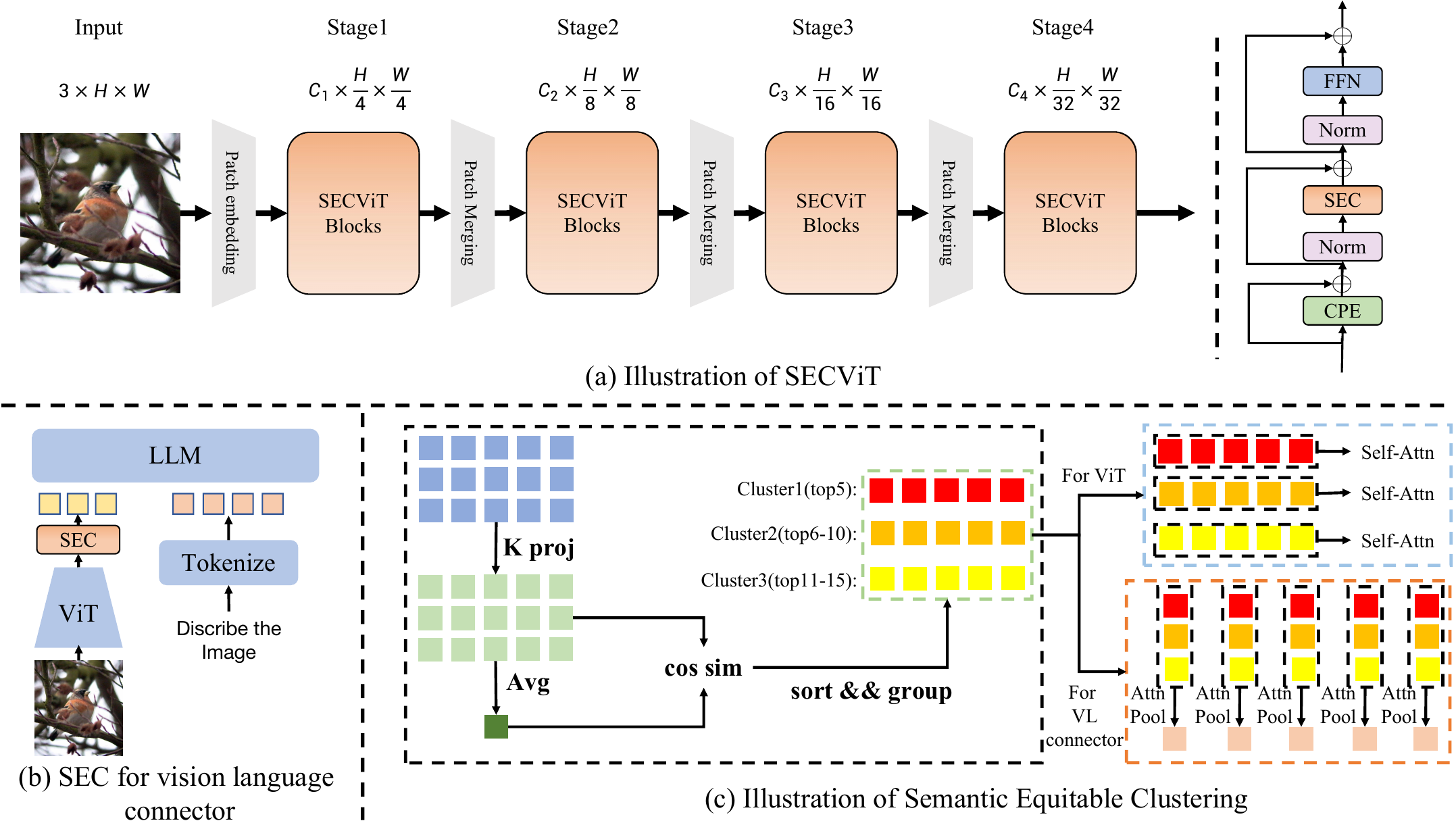}
    \vspace{-3mm}
    \caption{(a) Illustration of SECViT (b) Applying SEC to vision language connector. (c) Illustration of Semantic Equitable Clustering for ViT and vision language connector.}
    \label{fig:main}
    \vspace{-4mm}
\end{figure*}

%% file: sec/3_method.tex
\section{Method}
\label{sec:method}

\subsection{Overall Architecture}

The overall architecture of SECViT is shown in Fig.~\ref{fig:main}(a). SECViT consists of four stages with downsampling factors of $\frac{1}{4}$, $\frac{1}{8}$, $\frac{1}{16}$, and $\frac{1}{32}$, respectively. This structural design facilitates downstream tasks, such as object detection, in constructing feature pyramids. A SECViT block is composed of three modules. For each block, the input tensor $X_{in}\in \mathbb{R}^{C\times H\times W}$ is fed into the CPE to introduce the positional information. Then, The Self-Attention based on the Semantic Equitable Clustering (SEC) is employed to serve as the token mixer. The final FFN is utilized to integrate channel-wise information of tokens.

Beyond the design of the backbone, we also utilize SEC in the design of the vision language connector in MLLM~\citep{llava1.5}. For the vision tokens output by ViT, we use SEC to cluster the tokens. For each position corresponding to a cluster, we use attentive pooling to merge them into a single token, thereby reducing the number of vision tokens. The process is shown in Fig.~\ref{fig:main}(b).


\subsection{Semantic Equitable Clustering}

As previously mentioned, the design objectives of Semantic Equitable Clustering are threefold: \textbf{1)} Fully consider the semantic information contained in different tokens during clustering. \textbf{2)} Unlike k-means and other clustering methods that require multiple iterations, Semantic Equitable Clustering can complete clustering in a single step. \textbf{3)} Ensure an equal number of tokens in each cluster to facilitate parallel processing on GPUs. In the following paragraphs, we will describe in detail how our Semantic Equitable Clustering achieves these three objectives. And the whole process is illustrated in the Fig.~\ref{fig:main}(c).

\paragraph{Single Clustering Center Related to Semantics.}k-means is relatively complex for two reasons. \textbf{First}, it has multiple cluster centers, and each token needs to calculate its distance to each cluster center to determine its cluster membership. \textbf{Second}, the determination of each cluster center in k-means is not precise and requires multiple iterations to accurately establish the cluster centers. 

To address these two issues, we first discard the use of multiple cluster centers and instead calculate the distance between each token and a single center. Based on each token's distance to this center, we divide the tokens into different intervals. Then, to ensure that our chosen center contains the most comprehensive semantic information, we directly use the result of average pooling of all tokens as the center token. This is because, in most vision foundation models, the output of the average pool is assumed to contain the richest semantic information and is thus used for classification~\citep{SwinTransformer, cswin, CPVT, FAT}. Specifically, the process for determining the cluster center is shown in Eq.~\ref{eq:center}:
\begin{equation}
\label{eq:center}
\centering
\begin{split}
     Q = W_Q X, &K = W_K X, V = W_V X,\\
     &k_c = {\rm Pool}(K).
\end{split}
\end{equation}
Where $W_K$ is a learnable matrix. $k_c$ is the determined cluster center. $X$ is the set of input tokens.

\paragraph{Distance Metric Suitable for ViT.}Unlike the Euclidean distance calculation used in the k-means algorithm for computing the distance between tokens, during the actual computation of Self-Attention, similarity between query and key is computed through dot product. To better adapt to the characteristics of Self-Attention, we also measure the distance between tokens using a method similar to dot product. Specifically, we calculate the cosine similarity between the cluster center and each token, and then sort the tokens according to the magnitude of the computed results. The specific process is shown in Eq.~\ref{eq:sort}:
\begin{equation}
\label{eq:sort}
\centering
\begin{split}
     & sim = \frac{K \cdot k_c}{||K|| \cdot ||k_c||},\\
     & idx = {\rm argsort}(sim), \\
     Q^*=Q[idx]&, K^*=K[idx], V^*=V[idx].
\end{split}
\end{equation}
Where $sim$ is the similarity matrix between $K$ and $k_c$, the ${\rm argsort}(sim)$ returns the indices of $sim$ sorted in descending order. $Q^*,K^*,V^*$ are $Q,K,V$ rearranged according to ${\rm argsort}(sim)$.

\paragraph{Equally Partition Tokens based on Distance.}The obtained $Q^*$, $K^*$, and $V^*$ from the previous step have been sorted based on their distances to the cluster center. \textbf{For the design of vision backbone}, we directly group them, so tokens with similar distances to the cluster center are classified into the same cluster. This allows us to directly control an equal number of tokens in each cluster. This process can be clearly illustrated in Fig.~\ref{fig:main}(c) and denoted as follows:
\begin{equation}
\label{eq:part}
\centering
\begin{split}
Q_m=Q^*[{m\times N:\left( m+1 \right) N}], \\
K_m=K^{*}[{m\times N:\left( m+1 \right) N}], \\
V_m=V^{*}[{m\times N:\left( m+1 \right) N}].
\end{split}
\end{equation} where $N$ is the basic token number of each cluster for equal partition and $m$ is the index of the cluster

Based on the above steps, we have completed the clustering process that captures semantic information in the image with minimal sorting cost. Moreover, compared to k-means, we have achieved equi-partitioning of each cluster. After clustering is completed, we apply standard Self-Attention to the tokens within each cluster, thereby completing the interaction of information between tokens:
\begin{equation}
\label{eq:sca}
\centering
Y_m=\text{Attn}\left( Q_m,K_m,V_m \right).
\end{equation} 

\textbf{For the design of vision language connector, }we group the tokens according to their similarity, and the tokens within each group are interleaved, as shown in Eq.~\ref{eq:vlc}:
\begin{equation}
    \label{eq:vlc}
    \centering
    \begin{split}
        Q_{n}=Q^{*}[n:N:L], \\
        K_{n}=K^{*}[n:N:L], \\
        V_{n}=V^{*}[n:N:L].
    \end{split}
\end{equation}
in which $L$ is the token's sequence length, $n$ is the index of group tokens. $N$ is the basic token number of each cluster. After obtaining the token groups, we perform pooling on $Q$ to effectively reduce the number of tokens input to the LLM, with each group's output becoming a single token, as shown in Eq~\ref{eq:attnpool}.
\begin{equation}
    \label{eq:attnpool}
    Y_n={\rm Attn}({\rm Pool}(Q_{n}),K_{n},V_{n}).
\end{equation}

\subsection{Difference between SEC and EViT.}
We use the most representative example, EViT~\citep{evit}, to illustrate the differences between SEC and other methods based on the similarity between the global token and other tokens.

\textbf{Pruning v.s. Clustering.} Most similarity-based methods, such as EViT, are pruning methods, where tokens with low similarity to the [cls] token are merged during the forward process, thereby reducing the number of tokens and decreasing computational cost. In contrast, our proposed SECViT employs a clustering-based approach, performing attention operations within each cluster. 

\textbf{The role of the [cls] token.} In methods like EViT, the [cls] token serves as a measure of importance of a token. Each token computes its similarity to the [cls] token, with higher similarity tokens deemed more important. The less important tokens are abandoned. In contrast, in SEC, the [cls] token (obtained by average pooling over all tokens) measures similarity between tokens. Each token computes its similarity score to the [cls] token; tokens with similar scores are considered to be more similar and grouped into one cluster. Attention is calculated only within the same cluster.

%% file: sec/4_experiments.tex
\section{Experiments}
\label{sec:exp}
We first make strict comparison with hierarchical/plain baselines. Then we conduct experiments on a wide range of vision tasks for SECViT, including image classification, object detection, instance segmentation, and semantic segmentation. We also verify the role of SEC in MLLM based on LLaVA-1.5~\citep{llava1.5}. \textcolor{red}{More details, experiments, and comparison of models' efficiency} can be found in the \textcolor{red}{Appendix}.

\subsection{SEC for vision models}

\begin{table}[t]
\vspace{-3mm}
\setlength{\tabcolsep}{0.52mm}
    \centering
    \scalebox{0.77}{
    \begin{tabular}{c|c c c | c c c c}
    \toprule[1pt]
    Model & \makecell{Params\\(M)} & \makecell{FLOPs\\(G)} & \makecell{Throughput\\(imgs/s)} & Acc & $AP^b$ & $AP^m$ & mIoU \\
    \midrule[0.5pt]
    DeiT-S~\cite{deit} & 22 & 4.6 & 3204 & 79.8 & 44.5 & 40.1 & 43.0 \\
    \makecell{EViT-DeiT-S\\(keeprate=0.9)} & 22 & 4.0 & 3428 & 79.8 & \makecell{not\\suit} & \makecell{not\\suit} & \makecell{not\\suit} \\
    \rowcolor{gray!30}\makecell{SEC-DeiT-S\\(num\_cluster=4)} & 22 & 4.1 & 3412 & \makecell{80.5\\(\textcolor{red}{+0.7})} & \makecell{47.7\\(\textcolor{red}{+3.2})} & \makecell{42.7\\(\textcolor{red}{+2.6})} & \makecell{47.5\\(\textcolor{red}{+4.5})}\\
    \midrule[0.5pt]
    DeiT-B & 86 & 17.6 & 1502 & 81.8 & -- & -- & -- \\
    \rowcolor{gray!30}SEC-DeiT-B & 86 & 14.8 & 1682 & \makecell{82.4\\(\textcolor{red}{+0.6})} & -- & -- & -- \\
    \midrule[0.5pt]
    Swin-T & 29 & 4.5 & 1723 & 81.3 & 43.7 & 39.8 & 44.5 \\
    \rowcolor{gray!30}SEC-Swin-T & 29 & 4.8 & 1482 & \makecell{83.8\\(\textcolor{red}{+2.5})} & \makecell{48.3\\(\textcolor{red}{+4.6})} & \makecell{43.4\\(\textcolor{red}{+3.6})} & \makecell{49.3\\(+\textcolor{red}{4.8})} \\
    \midrule[0.5pt]
    Swin-S& 50 & 8.8 & 1006 & 83.0 & 45.7 & 41.1 & 47.6 \\
    \rowcolor{gray!30}SEC-Swin-S & 50 & 9.2 & 804 & \makecell{85.0\\(\textcolor{red}{+2.0})} & \makecell{50.2\\(\textcolor{red}{+4.5})} & \makecell{44.7\\(\textcolor{red}{+3.6})} & \makecell{51.3\\(\textcolor{red}{+3.7})} \\
    \midrule[0.5pt]
    \end{tabular}}
    \vspace{-3mm}
    \caption{Comparison with Hierarchy/Plain baselines. Inference speed are measured on the A100 GPU.}
    \vspace{-3mm}
    \label{tab:baseline}
\end{table}

\begin{table}[t]
    \centering
    \setlength{\tabcolsep}{0.8mm}
    \scalebox{0.77}{
    \begin{tabular}{c|c c|c c c}
        \toprule[1pt]
        Model & \makecell{Params\\(M)} & \makecell{FLOPs\\(G)} & Method &\makecell{Pretrain\\epoch} & Acc(\%) \\
        \midrule[0.5pt]
        Swin-B~\cite{SwinTransformer} & 88 & 15.4 & Supervised	& -- & 83.5 \\
        ConvNeXt V2-B~\cite{ConvNeXtV2} & 88 & 15.4 & Supervised & -- & 84.3 \\
        \rowcolor{gray!30}SEC-Swin-B & 88 & 16.2 & Supervised	&--	& 85.3 \\
        \midrule[0.5pt]
        Swin-B~\cite{SwinTransformer} & 88 & 15.4 & SimMIM~\cite{simmim} & 800 & 84.0(+0.5) \\
        ConvNeXt V2-B~\cite{ConvNeXtV2} & 88 & 15.4 &FCMAE~\cite{ConvNeXtV2} & 800 & 84.6(+0.3) \\
        \rowcolor{gray!30}SEC-Swin-B & 88 & 16.2 & SimMIM~\cite{simmim}	& 800 & 85.9(+0.6) \\
        \bottomrule[1pt]
    \end{tabular}}
    \vspace{-3mm}
    \caption{Comparison with baselines on self-supervised setting.}
    \vspace{-3mm}
    \label{tab:un}
\end{table}

\begin{table}[t]
    \centering
    \setlength{\tabcolsep}{2.9mm}
    \subfloat{
    \scalebox{0.75}{
    \begin{tabular}{c|c|c c|c}
        \toprule[1pt]
        \makecell{Cost} & Model & \makecell{Parmas\\(M)} & \makecell{FLOPs\\(G)} & \makecell{Top1-acc\\(\%)}\\
        \midrule[0.5pt]
        \multirow{12}{*}{\rotatebox{90}{\makecell{tiny model\\$\sim 2.5$G}}} 
        &PVTv2-b1~\cite{pvtv2} & 13 & 2.1 & 78.7 \\
        &TCFormer-light~\cite{tcformer} & 14 & 3.8 & 79.4 \\
        &QuadTree-B-b1~\cite{quadtree} & 14 & 2.3 & 80.0 \\
        &MPViT-XS~\cite{mpvit} & 11 & 2.9 & 80.9 \\
        &BiFormer-T~\cite{biformer} & 13 & 2.2 & 81.4 \\
        &CrossFormer-T~\cite{crossformer} & 28 & 2.9 & 81.5 \\
        &FAT-B2~\cite{FAT} & 14 & 2.0 & 81.9 \\
        &GC-ViT-XT~\cite{globalvit} & 20 & 2.6 & 82.0 \\
        &SMT-T~\cite{SMT} & 12 & 2.4 & 82.2 \\
        &RMT-T~\cite{fan2023rmt} & 14 & 2.5 & 82.4 \\
        &\cellcolor{gray!30}SECViT-T & \cellcolor{gray!30}15 & \cellcolor{gray!30}2.5 & \cellcolor{gray!30}\textbf{82.7} \\
        \midrule[0.5pt]
        \multirow{13}{*}{\rotatebox{90}{\makecell{small model\\$\sim 4.5$G}}} 
        &PS-ViT-B14~\cite{psvit} & 21 & 5.4 & 81.7 \\
        &DVT-T2T-ViT-19~\cite{dynamicvit2} & 39 & 6.2 & 81.9 \\
        &ConvNeXt-T~\cite{convnext} & 29 & 4.5 & 82.1 \\
        &TCFormer~\cite{tcformer} & 26 & 5.8 & 82.3 \\
        &SG-Former-S~\cite{sgformer} & 23 & 4.8 & 83.2 \\
        &StructViT-S-8-1~\cite{structvit} & 24 & 5.4 & 83.3 \\
        &InternImage-T~\cite{internimage} & 30 & 5.0 & 83.5 \\
        &MLLA-T~\cite{MLLA} & 25 & 4.2 & 83.5 \\
        &MaxViT-T~\cite{maxvit} & 31 & 5.6 & 83.6 \\
        &FAT-B3~\cite{FAT} & 29 & 4.4 & 83.6 \\
        &SMT-S~\cite{SMT} & 20 & 4.8 & 83.7 \\
        &BiFormer-S~\cite{biformer} & 26 & 4.5 & 83.8 \\
        &\cellcolor{gray!30}SECViT-S & \cellcolor{gray!30}27 & \cellcolor{gray!30}4.6 & \cellcolor{gray!30}\textbf{84.3} \\
        \midrule[0.5pt]
        \multirow{10}{*}{\rotatebox{90}{\makecell{base model\\$\sim 9.0$G}}}
        &ConvNeXt-S~\cite{convnext} & 50 & 8.7 & 83.1 \\
        &NAT-S~\cite{NAT} & 51 & 7.8 & 83.7 \\
        &Quadtree-B-b4~\cite{quadtree} & 64 & 11.5 & 84.0 \\
        &MOAT-1~\cite{MOAT} & 42 & 9.1 & 84.2 \\
        &InternImage-S~\cite{internimage} & 50 & 8.0 & 84.2 \\
        &GC-ViT-S~\cite{globalvit} & 51 & 8.5 & 84.3 \\
        &BiFormer-B~\cite{biformer} & 57 & 9.8 & 84.3 \\
        &iFormer-B~\cite{iformer} & 48 & 9.4 & 84.6 \\
        &FAT-B4~\cite{FAT} & 52 & 9.3 & 84.8 \\
        &\cellcolor{gray!30}SECViT-B & \cellcolor{gray!30}57 & \cellcolor{gray!30}9.8 & \cellcolor{gray!30}\textbf{85.2} \\
        \midrule[0.5pt]
        \multirow{9}{*}{\rotatebox{90}{\makecell{large model\\$\sim 18.0$G}}}
        &CrossFormer-L~\cite{crossformer} & 92 & 16.1 & 84.0 \\
        &SMT-L~\cite{SMT} & 81 & 17.7 & 84.6 \\
        &DaViT-B~\cite{davit} & 88 & 15.5 & 84.6 \\
        &SG-Former-B~\cite{sgformer} & 78 & 15.6 & 84.7 \\
        &iFormer-L~\cite{iformer} & 87 & 14.0 & 84.8 \\
        &InterImage-B~\cite{internimage} & 97 & 16.0 & 84.9 \\
        &GC-ViT-B~\cite{globalvit} & 90 & 14.8 & 85.0 \\
        &RMT-L~\cite{fan2023rmt} & 95 & 18.2 & 85.5 \\
        &\cellcolor{gray!30}SECViT-L & \cellcolor{gray!30}101 & \cellcolor{gray!30}18.2 & \cellcolor{gray!30}\textbf{85.7} \\
        \bottomrule[1pt]
    \end{tabular}}}
    \vspace{-3mm}
    \caption{Comparison with the state-of-the-art on ImageNet-1K classification.}
    \vspace{-3mm}
    \label{tab:ImageNet}
\end{table}

\begin{table*}[t]
\vspace{-3mm}
    \setlength{\tabcolsep}{1.72mm}
    \centering
    \scalebox{0.72}{
    \begin{tabular}{c|c c|c c c c c c|c c|c c c c c c}
        \toprule[1pt]
        \multirow{2}{*}{Backbone} & \multirow{2}{*}{\makecell{Params\\(M)}} & \multirow{2}{*}{\makecell{FLOPs\\(G)}} & \multicolumn{6}{c|}{Mask R-CNN $1\times$} & \multirow{2}{*}{\makecell{Params\\(M)}} & \multirow{2}{*}{\makecell{FLOPs\\(G)}} & \multicolumn{6}{c}{RetinaNet $1\times$}\\
         & & & $AP^b$ & $AP^b_{50}$ & $AP^b_{75}$ & $AP^m$ & $AP^m_{50}$ & $AP^m_{75}$ & & & $AP^b$ & $AP^b_{50}$ & $AP^b_{75}$ & $AP^b_S$ & $AP^b_{M}$ & $AP^b_{L}$ \\
         \midrule[0.5pt]
        PVTv2-B1~\cite{pvtv2} & 33 & 243 & 41.8 & 54.3 & 45.9 & 38.8 & 61.2 & 41.6 & 23 & 225 & 41.2 & 61.9 & 43.9 & 25.4 & 44.5 & 54.3 \\
        FAT-B2~\cite{FAT} & 33 & 215 & 45.2 & 67.9 & 49.0 & 41.3 & 64.6 & 44.0 & 23 & 196 & 44.0 & 65.2 & 47.2 & 27.5 & 47.7 & 58.8 \\
        RMT-T~\cite{fan2023rmt} & 33 & 218 & 47.1 &68.8 & 51.7 & 42.6 & 65.8 & 45.9 & 23 & 199 & 45.1 & 66.2 & 48.1 & 28.8 & 48.9 & 61.1 \\
        \rowcolor{gray!30}SECViT-T & 34 & 221 & \textbf{47.8} & \textbf{69.5} & \textbf{52.5} & \textbf{43.0} & \textbf{66.7} & \textbf{46.3} & 24 & 202 & \textbf{45.8} & \textbf{66.8} & \textbf{49.2} & \textbf{29.1} & \textbf{49.8} & \textbf{60.9} \\
        \midrule[0.5pt]
        MPViT-S~\cite{mpvit} & 43 & 268 & 46.4 & 68.6 & 51.2 & 42.4 & 65.6 & 45.7 & 32 & 248 & 45.7 & 57.3 & 48.8 & 28.7 & 49.7 & 59.2 \\
        MLLA-T~\cite{MLLA} & 44 & 255 & 46.8 & 69.5 & 51.5 & 42.1 & 66.4 & 45.0 & -- & -- & -- & -- & -- & -- & -- & -- \\
        STViT-S~\cite{stvit} & 44 & 252 & 47.6 & 70.0 & 52.3 & 43.1 & 66.8 & 46.5 & -- & -- & -- & -- & -- & -- & -- & -- \\
        RMT-S~\cite{fan2023rmt} & 46 & 262 & 49.0 & 70.8 & 53.9 & 43.9 & 67.8 & 47.4 & 36 & 244 & 47.8 & 69.1 & 51.8 & 32.1 & 51.8 & 63.5 \\
        \rowcolor{gray!30}SECViT-S & 45 & 262 & \textbf{49.9} & \textbf{70.9} & \textbf{54.7} & \textbf{44.6} & \textbf{68.3} & \textbf{47.7} & 35 & 240 & \textbf{48.4} & \textbf{69.4} & \textbf{52.0} & \textbf{31.3} & \textbf{53.3} & \textbf{63.8} \\
        \midrule[0.5pt]
        ScalableViT-B~\cite{ScalableViT} &95 & 349 & 46.8 & 68.7 & 51.5 & 42.5 & 65.8 & 45.9 & 85 & 330 & 45.8 & 67.3 & 49.2 & 29.9 & 49.5 & 61.0 \\
        InternImage-S~\cite{internimage} & 69 & 340 & 47.8 & 69.8 & 52.8 & 43.3 & 67.1 & 46.7 & -- & -- & -- & -- & -- & -- & -- & -- \\
        MLLA-S~\cite{MLLA} & 63 & 319 & 49.2 & 71.5 & 53.9 & 44.2 & 68.5 & 47.2 & -- & -- & -- & -- & -- & -- & -- & -- \\
        STViT-B~\cite{stvit} & 70 & 359 & 49.7 & 71.7 & 54.7 & 44.8 & 68.9 & 48.7 & -- & -- & -- & -- & -- & -- & -- & -- \\
        
        \rowcolor{gray!30}SECViT-B & 76 & 371 & \textbf{51.5} & \textbf{72.9} & \textbf{56.7} & \textbf{45.4} & \textbf{69.9} & \textbf{48.7} & 63 & 349 & \textbf{49.3} & \textbf{70.3} & \textbf{52.9} & \textbf{32.0} & \textbf{53.8} & \textbf{64.8} \\
        \midrule[0.5pt]
        Focal-B~\cite{focal} & 110 & 533 & 47.8 & 70.2 & 52.5 & 43.2 & 67.3 & 46.5 & 101 & 514 & 46.3 & 68.0 & 49.8 & 31.7 & 50.4 & 60.8 \\
        CSwin-B~\cite{cswin} & 97 & 526 & 48.7 & 70.4 & 53.9 & 43.9 & 67.8 & 47.3 & -- & -- & -- & -- & -- & -- & -- & -- \\
        InternImage-B~\cite{internimage} & 115 & 501 & 48.8 & 70.9 & 54.0 & 44.0 & 67.8 & 47.4 & -- & -- & -- & -- & -- & -- & -- & -- \\
        MLLA-B~\cite{MLLA} & 115 & 502 & 50.5 & 72.0 & 55.4 & 45.0 & 69.3 & 48.6 & -- & -- & -- & -- & -- & -- & -- & -- \\
        \rowcolor{gray!30}SECViT-L & 119 & 550 & \textbf{52.0} & \textbf{73.5} & \textbf{57.3} & \textbf{46.3} & \textbf{70.6} & \textbf{49.8} & 105 & 527 &  \textbf{50.2} & \textbf{71.4} & \textbf{53.9} & \textbf{33.2} & \textbf{54.5} & \textbf{66.3} \\
        \bottomrule[1pt]
    \end{tabular}}
    \vspace{-3mm}
    \caption{Comparison to other backbones using "$1\times$`` schedule on COCO.}
    \vspace{-3mm}
    \label{tab:COCO1x}
\end{table*}

\begin{table}[t]
    \vspace{-3mm}
    \centering
    \setlength{\tabcolsep}{0.7mm}
    \subfloat{
    \scalebox{0.72}{
    \begin{tabular}{c|c c|c c c c c c}
        \toprule[1pt]
         \multirow{2}{*}{Backbone} & \makecell{Params\\(M)} & \makecell{FLOPs\\(G)} & $AP^b$ & $AP^b_{50}$ & $AP^b_{75}$ & $AP^m$ & $AP^m_{50}$ & $AP^m_{75}$\\
          \midrule[0.5pt]
           \multicolumn{9}{c}{Mask R-CNN $3\times$+MS} \\
           \midrule[0.5pt]
          GC-ViT-T~\cite{globalvit} & 48 & 291 & 47.9 & 70.1 & 52.8 & 43.2 & 67.0 & 46.7 \\
          MLLA-T~\cite{MLLA} & 44 & 255 & 48.8 & 71.0 & 53.6 & 43.8 & 68.0 & 46.8 \\
          SMT-S~\cite{SMT} & 40 & 265 & 49.0 & 70.1 & 53.4 & 43.4 & 67.3 & 46.7\\
          InternImage-T~\cite{internimage} & 49 & 270 & 49.1 & 70.4 & 54.1 & 43.7 & 67.3 & 47.3 \\
          RMT-S~\cite{fan2023rmt} & 46 & 262 & 50.7 & 71.9 & 55.6 & 44.9 & 69.1 & 48.4\\
          \rowcolor{gray!30}SECViT-S & 45 & 262 & \textbf{51.6} & \textbf{72.5} & \textbf{55.9} & \textbf{45.6} & \textbf{69.9} & \textbf{48.8}\\
          \midrule[0.5pt]
          NAT-S~\cite{NAT} & 70 & 330 & 48.4 & 69.8 & 53.2 & 43.2 & 66.9 & 46.4 \\
          InternImage-S~\cite{internimage} & 69 & 340 & 49.7 & 71.1 & 54.5 & 44.5 & 68.5 & 47.8 \\
          SMT-B~\cite{SMT} & 52 & 328 & 49.8 & 71.0 & 54.4 & 44.0 & 68.0 & 47.3\\
          MLLA-S~\cite{MLLA} & 63 & 319 & 50.5 & 71.8 & 55.2 & 44.9 & 69.1 & 48.2 \\
          RMT-B~\cite{fan2023rmt} & 73 & 373 & 52.2 & 72.9 & 57.0 & 46.1 & 70.4 & 49.9 \\
          \rowcolor{gray!30}SECViT-B & 75 & 371 & \textbf{52.8} & \textbf{73.6} & \textbf{57.7} & \textbf{46.4} & \textbf{70.8} & \textbf{49.9}\\
          \midrule[0.5pt]
          \multicolumn{9}{c}{Cascade Mask R-CNN $3\times$+MS}\\
          \midrule[0.5pt]
          GC-ViT-T~\cite{globalvit} & 85 & 770 & 51.6 & 70.4 & 56.1 & 44.6 & 67.8 & 48.3 \\
          SMT-S~\cite{SMT} & 78 & 744 & 51.9 & 70.5 & 56.3 & 44.7 & 67.8 & 48.6 \\
          UniFormer-S~\cite{uniformer} & 79 & 747 & 52.1 & 71.1 & 56.6 & 45.2 & 68.3 & 48.9 \\
          RMT-S~\cite{fan2023rmt} & 83 & 741 & 53.2 & 72.0 & 57.8 & 46.1 & 69.8 & 49.8\\
          \rowcolor{gray!30}SECViT-S & 83 & 741 & \textbf{54.1} & \textbf{72.8} & \textbf{58.6} & \textbf{47.0} & \textbf{70.3} & \textbf{51.0}\\
          \midrule[0.5pt]
          NAT-S~\cite{NAT} & 108 & 809 & 51.9 & 70.4 & 56.2 & 44.9 & 68.2 & 48.6 \\
          GC-ViT-S~\cite{globalvit} & 108 & 866 & 52.4 & 71.0 & 57.1 & 45.4 & 68.5 & 49.3\\
          CSWin-S~\cite{cswin} & 92 & 820 & 53.7 & 72.2 & 58.4 & 46.4 & 69.6 & 50.6 \\
          UniFormer-B~\cite{uniformer} & 107 & 878 & 53.8 & 72.8 & 58.5 & 46.4 & 69.9 & 50.4 \\
          RMT-B~\cite{fan2023rmt} & 111 & 852 & 54.5 & 72.8 & 59.0 & 47.2 & 70.5 & 51.4 \\
          \rowcolor{gray!30}SECViT-B & 114 & 849 & \textbf{55.4} & \textbf{74.1} & \textbf{59.9} & \textbf{47.8} & \textbf{71.7} & \textbf{51.7} \\
          \bottomrule
    \end{tabular}}}
    
    \vspace{-3mm}
    \caption{Comparison with other backbones using "$3\times+\mathrm{MS}$`` schedule on COCO.}
    \label{tab:COCO3x}
    \vspace{-5mm}
\end{table}

\begin{table}[ht]
    \centering
    \setlength{\tabcolsep}{0.45mm}
    \scalebox{0.8}{
    \begin{tabular}{c|c c c|c c c}
    \toprule[1pt]
    \multirow{3}{*}{Model} & \multicolumn{3}{c|}{Semantic FPN 80K} & \multicolumn{3}{c}{Upernet 160K} \\
    & \makecell{Params\\(M)} & \makecell{FLOPs\\(G)} & \makecell{mIoU\\(\%)} & \makecell{Params\\(M)} & \makecell{FLOPs\\(G)} & \makecell{mIoU$_{ss}$\\(\%)} \\
    \midrule[0.5pt]
    PVTv2-B1~\cite{pvtv2} & 18 & 136 & 42.5 & -- & -- & -- \\
    VAN-B1~\cite{VAN} & 18 & 140 & 42.9 & -- & -- & -- \\
    RMT-T~\cite{fan2023rmt} & 17 & 136 & 46.4 & -- & -- & -- \\
    \rowcolor{gray!30}SECViT-T & 18 & 136 & \textbf{47.2} & 44 & 894 & \textbf{48.8} \\
    \midrule[0.5pt]
    StructViT-S~\cite{structvit} & 26 & 271 & 46.9 & -- & -- & -- \\
    MogaNet-S~\cite{iclr2024MogaNet} & 29 & 189 & 47.7 & 55 & 946 & 49.2 \\
    SMT-S~\cite{SMT} & -- & -- & -- & 50 & 935 & 49.2 \\
    SGFormer-S~\cite{sgformer} & 25 & 205 & 49.0 & 52.5 & 989 & 49.9 \\
    RMT-S~\cite{fan2023rmt} & 30 & 180 & 49.4 & 56 & 937 & 49.8 \\
    \rowcolor{gray!30}SECViT-S & 30 & 180 & \textbf{49.6} & 56 & 936 & \textbf{50.6} \\
    \midrule[0.5pt]
    MogaNet-B~\cite{iclr2024MogaNet} & -- & -- & -- & 74 & 1050 & 50.1 \\
    InterImage-S~\cite{internimage} & -- & -- & -- & 80 & 1017 & 50.2 \\
    StructViT-B~\cite{structvit} & 54 & 529 & 48.5 & -- & -- & -- \\
    RMT-B~\cite{fan2023rmt} & 57 & 294 & 50.4 & 83 & 1051 & 52.0 \\
    \rowcolor{gray!30}SECViT-B & 60 & 291 & \textbf{50.7} & 86 & 1048 & \textbf{52.2} \\
    \midrule[0.5pt]
    MogaNet-L~\cite{iclr2024MogaNet} & -- & -- & -- & 113 &1176 & 50.9 \\
    MLLA-B~\cite{MLLA} & -- & -- & -- & 128 & 1183 & 51.9 \\
    SGFormer-B~\cite{sgformer} & 81 & 475 & 50.6 & 109 & 1304 & 52.0 \\
    RMT-L~\cite{fan2023rmt} & 98 & 482 & 51.4 & 125 & 1241 & 52.8 \\
    \rowcolor{gray!30}SECViT-L & 103 & 475 & \textbf{52.2} & 131 & 1256 & \textbf{53.8} \\
    \bottomrule[1pt]
    \end{tabular}}
    \vspace{-3mm}
    \caption{ Comparison with the state-of-the-art on ADE20K.}
    \vspace{-3mm}
    \label{tab:ade20k}
\end{table}

\begin{table*}[t]
    \centering
    \vspace{-3mm}
    \setlength{\tabcolsep}{4.5mm}
    \scalebox{0.7}{
    \begin{tabular}{c c c c c c c c c c c}
    \toprule[1pt]
         Model &  Connector & V-T Num & Time & Speed & TextVQA & GQA & VQAv2 & POPE & MME \\
         \midrule[0.5pt]
         LLaVA-1.5 &  MLP & 576+1 & 194s & $1.0\times$ & 58.2 & 62.0 & 78.5 & 86.1 & 1510.7 \\
         \midrule[0.5pt]
         LLaVA-1.5+Resampler & Resampler & 288+1 & 126s & $1.5\times$ & 52.1 & 56.8 & 76.0 & 83.1 & 1393.2 \\
         LLaVA-1.5+EViT & MLP+EViT & 288+1 & 126s & $1.5\times$ & 54.6 & 60.0 & 77.9 & 84.3 & 1483.2 \\
         \rowcolor{gray!30}LLaVA-1.5+SEC & MLP+SEC & 288+1 & 126s & \textbf{$1.5\times$} & \textbf{60.1} & \textbf{63.5} & \textbf{78.9} & \textbf{87.7} & \textbf{1510.7} \\
         \midrule[0.5pt]
         LLaVA-1.5+Resampler & Resampler & 256+1 & 116s & $1.7\times$ & 51.6 & 56.0 & 75.2 & 82.7 & 1387.2 \\
         LLaVA-1.5+Pool & MLP+Pool & 256+1 & 116s & $1.7\times$ & 52.4 & 57.6 & 76.4 & 83.3 & 1415.5 \\
         LLaVA-1.5+EViT & MLP+EViT & 256+1 & 116s & $1.7\times$ & 52.8 & 59.6 & 77.1 & 83.7 & 1443.7 \\
         \rowcolor{gray!30}LLaVA-1.5+SEC & MLP+SEC & 256+1 & 116s & \textbf{$1.7\times$} & \textbf{59.6} & \textbf{63.2} & \textbf{78.6} & \textbf{87.1} & \textbf{1505.2} \\
         \midrule[0.5pt]
         LLaVA-1.5+Resampler & Resampler & 192+1 & 102s & $1.9\times$& 50.1 & 55.2 & 74.3 & 82.7 & 1337.6 \\
         LLaVA-1.5+EViT & MLP+EViT & 192+1 & 102s & $1.9\times$& 51.6 & 58.6 & 76.3 & 83.1 & 1427.6 \\
         \rowcolor{gray!30}LLaVA-1.5+SEC & MLP+SEC & 192+1 & 102s &\textbf{$1.9\times$} & \textbf{57.7} & \textbf{62.7} & \textbf{78.4} & \textbf{86.7} & \textbf{1500.1} \\
         \midrule[0.5pt]
         LLaVA-1.5+Resampler & Resampler & 144+1 & 94s & $2.1\times$ & 47.6 & 54.6 & 72.0 & 81.9 & 1293.7 \\
         LLaVA-1.5+Pool & MLP+Pool & 144+1 & 94s & $2.1\times$ & 50.0 & 56.2 & 73.6 & 81.9 & 1310.7 \\
         LLaVA-1.5+EViT & MLP+EViT & 144+1 & 94s & $2.1\times$ & 51.2 & 58.0 & 76.0 & 83.1 & 1393.6 \\
         \rowcolor{gray!30}LLaVA-1.5+SEC & MLP+SEC & 144+1 & 94s & \textbf{$2.1\times$} & \textbf{56.8} & \textbf{62.0} & \textbf{78.0} & \textbf{86.1} & \textbf{1487.1} \\
         \bottomrule[1pt]
    \end{tabular}}
    \vspace{-3mm}
    \caption{Comparison of different vision language connectors on LLaVA-1.5. ``V-T Num" denotes the quantity of visual tokens. The computation expense is impacted by V-T Num, with larger values resulting in higher costs. ``Speed" refers to the comparative inference velocity relative to LLaVA-1.5. ``Time" is the average inference time. Inference speed are measured on the A100.}
    \vspace{-3mm}
    \label{tab:MLLM}
\end{table*}

\begin{table*}[t]
\centering
\setlength{\tabcolsep}{3.2mm}
\scalebox{0.7}{
\begin{tabular}{l|c|c|c|c|ccccc|c}
\toprule
Model  & LLM &Connector &V-T Num &Res & TextVQA& GQA& VQAv2& VisWiz& SQA$_{img}$ & Speed ($\uparrow$)\\
\midrule
\multicolumn{11}{l}{7B LLM}\\

Shikra~\citep{chen2023shikra}              & Vicuna-7B & MLP       & 257 & 224& -    & -    & 77.4 & -    & -  & -\\
IDEFICS-9B~\citep{laurenccon2024obelics}   & LLaMA-7B  & Cross Attn   & 257 & 224& -    & 38.4 & 50.9 & 35.5 & -  & -\\
Qwen-VL~\citep{bai2023qwenvl}               & Qwen-7B   & Resampler    & 256 & 448& -    & 59.3 & 78.8 & 35.2 & 67.1 & -\\
Qwen-VL-Chat~\citep{bai2023qwenvl}          & Qwen-7B   & Resampler    & 256 & 448& -    & 57.5 & 78.2 & 38.9 & 68.2 & -\\
\midrule
LLaVA-1.5~\citep{liu2023improvedllava}     & Vicuna-7B & MLP       & 577 & 336& 58.2 & 62.0 & 78.5 & 50.0 & 66.8 & $1.0\times$\\
\midrule
\rowcolor{gray!30}LLaVA-1.5+SEC (ours) & Vicuna-7B & MLP+SEC       & 257 & 336& \textbf{59.6} & \textbf{63.2}& \textbf{78.9} & \textbf{52.8} & \textbf{69.6} & $\mathbf{1.7\times}$\\
\midrule

\multicolumn{11}{l}{13B LLM}\\

InstructBLIP~\citep{dai2023instructblip}    & Vicuna-13B & Q-Former     & 32  & 224& -    & 49.5  & -    & 33.4 & 63.1 & -\\
BLIP-2~\citep{li2023blip}                   & Vicuna-13B & Q-Former     & 32  & 224& -    & 41.0  & 41.0 & 19.5 & 61.0 & -\\
\midrule
LLaVA-1.5~\citep{liu2023improvedllava}     & Vicuna-13B & MLP       & 577 & 336& 61.2 & 63.3  & \textbf{80.0} & 53.6 & 71.6 & $1.0\times$\\
\midrule
\rowcolor{gray!30}LLaVA1.5+SEC (ours)& Vicuna-13B & MLP+SEC       & 257 & 336& \textbf{62.3} & \textbf{64.3} & \textbf{80.0} & \textbf{54.7} & \textbf{72.0} & $\mathbf{1.8\times}$\\
\bottomrule
\end{tabular}}
\vspace{-3mm}
\caption{Results on General VQA tasks.}
\vspace{-3mm}
\label{tab:vqa}
\end{table*}
\begin{table*}[t]
\centering
\setlength{\tabcolsep}{5.2mm}
\scalebox{0.7}{
\begin{tabular}{l|c|c|c|c|ccc|c}
\toprule
Model  & LLM &Connector &V-T Num &Res & POPE& MMB& MM-Vet& Speed ($\uparrow$)\\
\midrule

\multicolumn{9}{l}{7B LLM}\\

MiniGPT-4~\citep{zhu2023minigpt4}         & Vicuna-7B & Resampler    & 32  & 224& 72.2  & 24.3 & 22.1& -\\
mPLUG-Owl2~\citep{ye2024mplugowl}           & LLaMA2-7B & Resampler    & 32  & 224& -    & 49.4 & -   & -\\
\midrule
LLaMA-AdapterV2~\citep{gao2023llamaadapter}& LLaMA2-7B & LLaMA-Adapter& 257 & 224& -    &  41.0 & 31.4& -\\
Shikra~\citep{chen2023shikra}              & Vicuna-7B & MLP       & 257 & 224& -    & 58.8 & -   & -\\
Qwen-VL~\citep{bai2023qwenvl}               & Qwen-7B   & Resampler    & 256 & 448& -     & 38.2 & -   & -\\
Qwen-VL-Chat~\citep{bai2023qwenvl}          & Qwen-7B   & Resampler    & 256 & 448& -    & 60.6 & -   & -\\
\midrule
LLaVA-1.5~\citep{liu2023improvedllava}     & Vicuna-7B & MLP       & 577 & 336& 86.1 & 64.3 & 31.1& $1.0\times$\\
\midrule
\rowcolor{gray!30}LLaVA1.5+SEC (ours)& Vicuna-7B & MLP+SEC       & 145 & 336& \textbf{86.1} & \textbf{68.4} & \textbf{31.7}& $\mathbf{2.1\times}$\\
\midrule

\multicolumn{9}{l}{13B LLM}\\

MiniGPT-4~\citep{zhu2023minigpt4}         & Vicuna-13B & Resampler    & 32  & 224& -    & -    & 24.4& -\\
BLIP-2~\citep{li2023blip}                   & Vicuna-13B & Q-Former     & 32  & 224& 85.3 & -    & 22.4& -\\
\midrule
LLaVA-1.5~\citep{liu2023improvedllava}     & Vicuna-13B & MLP       & 577 & 336& 86.2 & 67.7 & 36.1& $1.0\times$\\
\midrule
\rowcolor{gray!30}LLaVA-1.5+SEC (ours)& Vicuna-13B & MLP+SEC & 145 & 336& \textbf{86.4} & \textbf{69.2} & \textbf{37.3}& $\mathbf{2.2\times}$\\
\bottomrule
\end{tabular}}
\vspace{-3mm}
\caption{Results on benchmark designed for MLLMs.}
\vspace{-5mm}
\label{tab:mllmbench}
\end{table*}

\paragraph{Strict Comparison with Baselines.}We select two baselines: hierarchical backbone Swin-Transformer~\citep{SwinTransformer} and plain backbone DeiT~\citep{deit} to make a comparison with our SEC based model. In the comparison models (SEC-Swin and SEC-DeiT), we merely substitute the attention mechanism in the original model with our SEC based Self-Attention and without introducing any other modules. As shown in Tab.~\ref{tab:baseline}, we conduct experiments on image classification, object detection, insatance segmentation and semantic segmentation, the simple replacement of the attention mechanism yields significant advantages in both performance and efficiency.

In addition to the supervised scenario, we also train the model with SimMIM~\citep{simmim} in the self-supervised scenario. As shown in Tab.~\ref{tab:un}, SEC also performs exceptionally well in the self-supervised scenario.

\paragraph{Image Classification.}We compare our SECViT with numerous state-of-the-art models, the results are shown in Tab.\ref{tab:ImageNet}. We adopt the training strategy proposed in DeiT~\citep{deit}, with the only supervision is cross entropy loss. All of our models are trained from scratch for 300 epochs with the input resolution of $224\times 224$. SECViT consistently outperforms preceding models across all scales. Notably, SECViT-S attains a Top1-accuracy of \textbf{84.3\%} with a mere \textbf{27M} parameters and \textbf{4.6G} FLOPs. The comparison of the models' efficiency can be found in \textcolor{red}{Appendix}.

\paragraph{Object Detection and Instance Segmentation. }We utilize MMDetection~\citep{mmdetection} to implement Mask-RCNN~\citep{maskrcnn}, Cascade Mask R-CNN~\citep{cai18cascadercnn}, and RetinaNet~\citep{retinanet} to evaluate the performance of the SECViT. Tab.~\ref{tab:COCO3x} and Tab.~\ref{tab:COCO1x} show the results of SECViT with different detection frameworks. The results show that SECViT performs better than its counterparts in all comparisons.

\paragraph{Semantic Segmentation.}We utilize Semantic FPN~\citep{semanticfpn} and UperNet~\citep{upernet} to validate our SECViT's performance, implementing these frameworks via MMSegmentation~\citep{mmsegmentation}. The results of semantic segmentation can be found in the Tab.~\ref{tab:ade20k}. All the FLOPs are measured with the input resolution of $512\times2048$.  SECViT achieves the best performance in all settings.

\subsection{SEC for MLLM}
SEC can greatly facilitate the design of vision language connectors in MLLMs. First, we conduct a rigorous comparison between SEC and various baseline vision language connectors based on LLaVA-1.5. Then, we compare LLaVA-1.5+SEC with several popular contemporary MLLMs.

\paragraph{Strict Comparison with Baselines.}In Tab.~\ref{tab:MLLM}, we strictly compare various commonly used vision language connectors, including MLP, Resampler~\citep{bai2023qwenvl}, Pooling, and EViT~\citep{evit}, which has achieved success in the design of ViT. Among these, MLP is the original design in LLaVA-1.5~\citep{llava1.5}, capable of achieving good results. However, it incurs significant computational cost due to the excessive vision tokens. To address this issue, some connectors attempt to use fewer vision tokens to accelerate LLaVA-1.5. Nonetheless, these adjustments inevitably lead to performance degradation. The results in Tab.~\ref{tab:MLLM} show that using SEC can effectively accelerate the inference of LLaVA-1.5 without causing performance degradation, and can even improve the performance of LLaVA-1.5 to a certain extent.

\paragraph{Comparison with Popular MLLMs.}In Tab.~\ref{tab:vqa} and Tab.~\ref{tab:mllmbench}, we compare LLaVA-1.5 equipped with SEC as a vision-language connector with other MLLMs. It is evident that SEC not only enhances the performance of MLLMs across various benchmarks but also significantly improves the efficiency of the models. This fully demonstrates the effectiveness of SEC in extracting visual information.

\subsection{Ablation Study}
In this section, we present some of the ablation study results for SEC, and more results can be found in the \textcolor{red}{Appendix}.

\paragraph{Number of Vision Tokens in Each Clusters.}The number of vision tokens has a significant impact on the performance and speed of the model. We thoroughly investigate the effect of the number of vision tokens on SECViT. As shown in Tab.~\ref{tab:ab_num_v}, the number of vision tokens in each cluster greatly influences the model's performance. Specifically, in downstream dense prediction tasks, having too few tokens in each cluster leads to substantial performance degradation. When the number of tokens in each cluster is too large, the model's performance does not see a significant improvement, but its speed decreases.

\begin{table}[ht]
    \centering
    \setlength{\tabcolsep}{0.52mm}
    \scalebox{0.8}{
    \begin{tabular}{c|c c c|c c c c}
    \toprule[1pt]
    V-T num & \makecell{Params\\(M)} & \makecell{FLOPs\\(G)} & \makecell{Throughput\\(imgs/s)} & Acc & $AP^b$ & $AP^m$ & mIoU \\
    \midrule[0.5pt]
    98 & 15 & 2.5 & 2004 & 82.7 & 47.8 & 43.0 & 47.2 \\
    \midrule[0.5pt]
    196 & 15 & 3.1 & 1722 & \makecell{83.0\\(\textcolor{red}{+0.3})} & \makecell{48.2\\(\textcolor{red}{+0.4})} & \makecell{43.4\\(\textcolor{red}{+0.4})} & \makecell{47.5\\(\textcolor{red}{+0.3})} \\
    64 & 15 & 2.5 & 1946 & \makecell{82.7\\(\textcolor{red}{+0.0})} & \makecell{47.8\\(\textcolor{red}{+0.0})} & \makecell{42.8\\(\textcolor{blue}{-0.2})} & \makecell{46.9\\(\textcolor{blue}{-0.3})}\\
    49 & 15 & 2.4 & 2102 & \makecell{82.6\\(\textcolor{blue}{-0.1})} & \makecell{47.5\\(\textcolor{blue}{-0.3})} & \makecell{42.7\\(\textcolor{blue}{-0.3})} & \makecell{47.7\\(\textcolor{blue}{-0.5})} \\
    24 & 15 & 2.3 & 2186 & \makecell{82.0\\(\textcolor{blue}{-0.7})} & \makecell{45.9\\(\textcolor{blue}{-1.9})} & \makecell{40.6\\(\textcolor{blue}{-2.4})} & \makecell{44.6\\(\textcolor{blue}{-2.6})} \\
    \bottomrule[1pt]
    \end{tabular}}
    \vspace{-3mm}
    \caption{Effect of the number of vision tokens in each cluster. ``V-T num" means the number of vision tokens in each cluster. The experiments are conducted based on SECViT-T.}
    \vspace{-3mm}
    \label{tab:ab_num_v}
\end{table}

\subsection{Visualization of SEC}
\begin{figure}[t]
    \centering
    \includegraphics[width=0.99\linewidth]{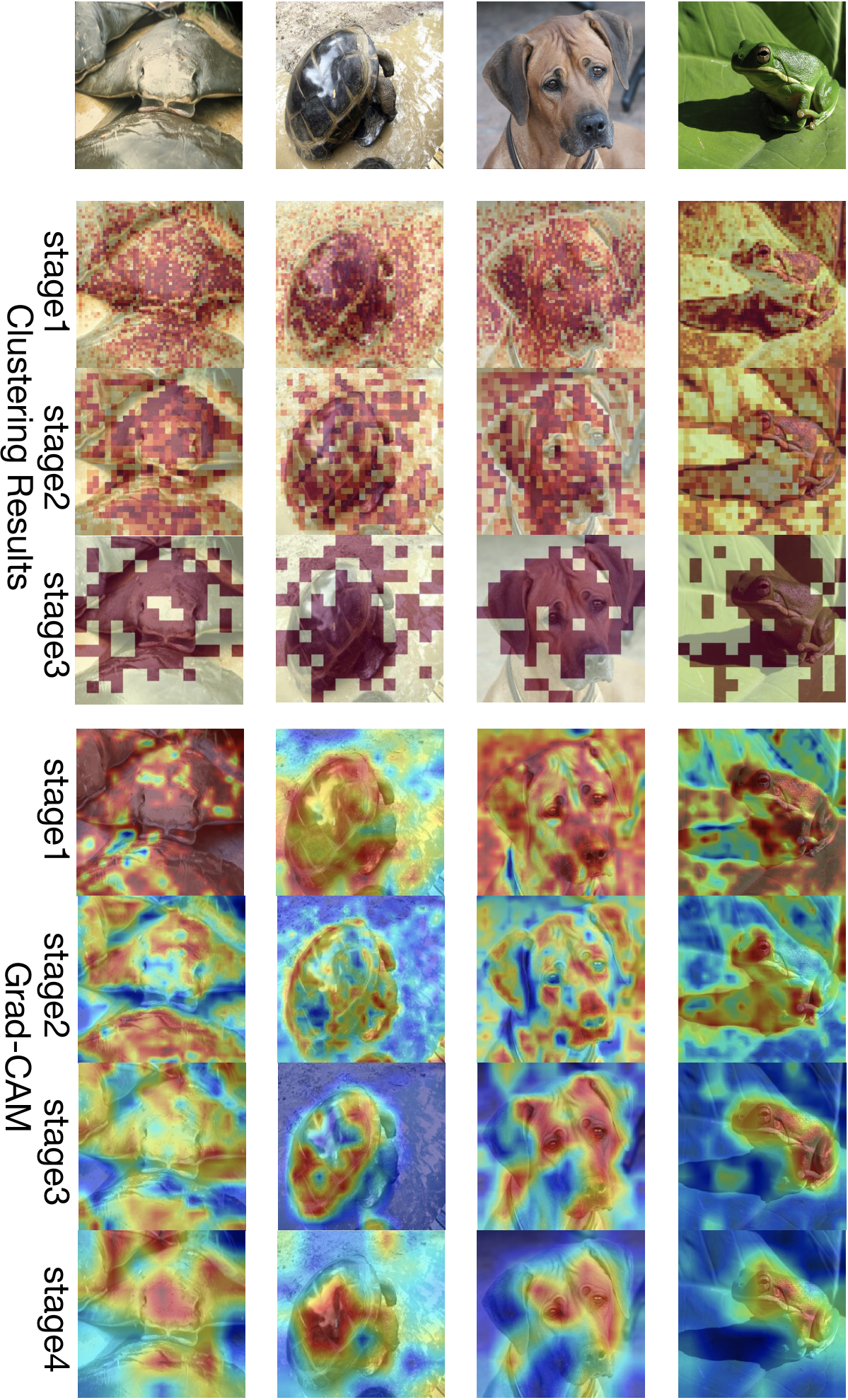}
    \vspace{-3mm}
    \caption{Visualization for SEC. }
    \vspace{-3mm}
    \label{fig:visualization}
\end{figure}

To further understand the working mechanism of SEC, we visualize some clustering results for SECViT. As shown in Fig.~\ref{fig:visualization}, the left side presents the clustering results of vision tokens at different stages of the model. From the clustering results, we analyze that in the shallow layers, the model distinguishes fine-grained features well, while in the deeper layers, it captures global semantic features effectively. The right side shows the Grad-CAM diagrams at different stages of the model, from which we can draw similar conclusions to the clustering results. More visualization results can be found in \textcolor{red}{Appendix}.

\paragraph{Number of Vision Tokens Outputs by SEC.}
MLLM is quite sensitive to the number of vision tokens. We conduct a detailed exploration based on LLaVA-1.5 regarding the number of vision tokens output by SEC, as shown in Tab.~\ref{tab:ab_num_mllm}. The first row represents the speed and performance of the original LLaVA-1.5 without using SEC. Compared to LLaVA-1.5, employing SEC effectively reduces the number of vision tokens and improves training efficiency. As the number of vision tokens decreases, the model's performance shows a slight decline, but its efficiency is further enhanced.

\begin{table}[t]
    \centering
    \setlength{\tabcolsep}{0.4mm}
    \scalebox{0.75}{
    \begin{tabular}{c|cc|ccccc}
    \toprule[1pt]
    V-T num & Time & Speed & TextVQA & GQA & VQAv2 & POPE & MM-Vet  \\
    \midrule[0.5pt]
    576+1 & 21h & $1.0\times$ & 58.2 & 62.0 & 78.5 & 86.1 & 31.1  \\
    \midrule[0.5pt]
    288+1 & \textcolor{red}{14h} & \textcolor{red}{$1.5\times$} & 60.1(\textcolor{red}{+1.9}) & 63.5(\textcolor{red}{+1.5}) & 78.9(\textcolor{red}{+0.4})& 87.7(\textcolor{red}{+1.6}) & 33.2(\textcolor{red}{+2.1}) \\
    256+1 & \textcolor{red}{13h} & \textcolor{red}{$1.6\times$} & 59.6(\textcolor{red}{+1.4}) & 63.2(\textcolor{red}{+0.3}) & 78.6(\textcolor{red}{+0.1}) & 87.1(\textcolor{red}{+1.0}) & 32.7(\textcolor{red}{+1.6}) \\
    192+1 & \textcolor{red}{11h} & \textcolor{red}{$1.9\times$} & 57.7(\textcolor{blue}{-0.5}) & 62.7(\textcolor{red}{+0.7}) & 78.4(\textcolor{blue}{-0.1}) & 86.7(\textcolor{red}{+0.6}) & 32.1(\textcolor{red}{+1.0}) \\
    144+1 & \textcolor{red}{10h} & \textcolor{red}{$2.1\times$} & 56.8(\textcolor{blue}{-1.4}) & 62.0(\textcolor{red}{+0.0}) & 78.0(\textcolor{blue}{-0.5}) & 86.1(\textcolor{red}{+0.0}) & 31.7(\textcolor{red}{+0.6}) \\
    \bottomrule[1pt]
    \end{tabular}}
    \vspace{-3mm}
    \caption{Effect of the number of vision tokens outputs by SEC. ``V-T num" means the number of vision tokens output by SEC. The experiments are conducted based on LLaVA-1.5~\citep{llava1.5}.}
    \vspace{-3mm}
    \label{tab:ab_num_mllm}
\end{table}

%% file: sec/5_conclusion.tex
\section{Conclusion}
\label{sec:conclusion}
We propose a simple and straightforward clustering method for vision tokens—Semantic Equitable Clustering (SEC). This method assigns each token to a cluster by calculating the similarity between each token and a global token, and completes the whole clustering process in only one step. Our clustering method takes into account the semantic information contained in the tokens, and ensures an equal number of tokens in each cluster, facilitating efficient parallel processing on modern GPUs. Based on Semantic Equitable Clustering, we designed SECViT, a versatile vision backbone that achieves impressive results across various vision tasks, including image classification, object detection, instance segmentation, and semantic segmentation. Besides, SEC can also be conveniently applied to multimodal large language models (MLLM) to serve as a vision language connector and benefits the model's efficiency.

%% file: sec/6_ack.tex
\section{Acknowledgements}
This work is partially funded by Beijing Natural Science Foundation (4252054), Youth Innovation Promotion Association CAS(Grant No.2022132), Beijing Nova Program(20230484276), and CCF-Kuaishou Large Model Explorer Fund (NO. CCF-KuaiShou 2024005).

%% file: sec/X_suppl.tex
\clearpage
\setcounter{page}{1}
\maketitlesupplementary
\begin{appendices}
\section{Experimental Details}

\paragraph{SECViT's Architectures.}SECViT's architecture details are illustrated in Table~\ref{tab:arch}. In SECViT, we adopt four $3\times 3$ convolutions to embed the input image into tokens, batch normalization and GELU are used after each convolution.
$3\times 3$ convolutions with stride 2 are used between stages to reduce the feature resolution.
$3\times 3$ DWConvs are adopted in CPE. For all models, we set the number of clusters in the first three stages to 32, 8, and 2, respectively.

\begin{table*}[t]
    \centering
    \setlength{\tabcolsep}{0.8mm}
    \begin{tabular}{c|c c c c|c c}
         \toprule[1pt]
         Model & Blocks & Channels & Heads & Ratios & Params(M) & FLOPs(G)\\
         \midrule[0.5pt]
         SECViT-T & [2, 2, 9, 2] & [64, 128, 256, 512] & [2, 4, 8, 16] & 3 & 15 & 2.5 \\
         SECViT-S & [4, 4, 18, 4] & [64, 128, 256, 512] & [2, 4, 8, 16] & 3 & 27 & 4.6 \\
         SECViT-B & [4, 8, 26, 9] & [80, 160, 320, 512] & [2, 4, 8, 16] & 3 & 57 & 9.8 \\
         SECViT-L & [4, 8, 26, 9] & [112, 224, 448, 640] & [4, 8, 14, 20] & 3 & 101 & 18.2 \\
         SECViT-XL & [6, 12, 28, 12] & [128, 256, 512, 1024] & [4, 8, 16, 32] & 3 & 205 & 36.4 \\
         \bottomrule[1pt]
    \end{tabular}
    \vspace{-3mm}
    \caption{Detailed Architectures of our models.}
    \vspace{-3mm}
    \label{tab:arch}
\end{table*}

\section{More experimental Results}
\paragraph{Efficiency Comparison.}In Tab.~\ref{tab:efficiency}, we compare the inference efficiency of various models in detail. From this, we can see that the ViT based on SEC demonstrates the best performance-speed tradeoff.
\begin{table*}[ht]
    \centering
    \setlength{\tabcolsep}{1.9mm}
    \scalebox{0.9}{
    \begin{tabular}{c|c c c|c}
    \toprule[1pt]
        Model & Params(M) & FLOPs(G) & Throughput(imgs/s) & Top1-Acc(\%) \\
        \midrule[0.5pt]
        DeiT-S~\cite{deit} & 22 & 4.6 & 3204 & 79.8 \\
        \makecell{EViT-DeiT-S (keeprate=0.9)~\cite{evit}} & 22 & 4.0 & 3428 & 79.8 \\
        \rowcolor{gray!30}SEC-DeiT-S (num\_cluster=4) & 22 & 4.1 & 3412 & 80.5 \\
        \midrule[0.5pt]
        DeiT-B~\cite{deit} & 86 & 17.6 & 1502 & 81.8 \\
        \rowcolor{gray!30}SEC-DeiT-B & 86 & 14.8 & 1682 & 82.4 \\
        \midrule[0.5pt]
        PVTv2-b1~\cite{pvtv2} & 13 & 2.1 & 2204 & 78.7 \\
        TCFormer-light~\cite{tcformer} & 14 & 3.8 & 417 & 79.4 \\
        MPViT-XS~\cite{mpvit} & 11 & 2.9 & 1496 & 80.9 \\
        BiFormer-T~\cite{biformer} & 13 & 2.2 & 1634 & 81.4 \\
        CMT-XS~\cite{cmt} & 15 & 1.5 & 1476 & 81.8 \\
        GC-ViT-XT~\cite{globalvit} & 20 & 2.6 & 1308 & 82.0 \\
        SMT-T~\cite{SMT} & 12 & 2.4 & 638 & 82.2 \\
        RMT-T~\cite{fan2023rmt} & 14 & 2.5 & 1438 & 82.4 \\
        \rowcolor{gray!30}SECViT-T & 15 & 2.5 & 2004 & 82.7 \\
        \midrule[0.5pt]
        Swin-T~\cite{SwinTransformer} & 29 & 4.5 & 1723 & 81.3 \\
        PS-ViT-B14~\cite{psvit} & 21 & 5.4 & 1986 & 81.7 \\
        DVT-T2T-ViT-19~\cite{dynamicvit2} & 39 & 6.2 & 1268 & 81.9 \\
        SGFormer-S~\cite{sgformer} & 23 & 4.8 & 952 & 83.2 \\
        CMT-S~\cite{cmt} & 25 & 4.0 & 846 & 83.5 \\
        CSwin-S~\cite{cswin} & 35 & 6.9 & 972 & 83.6 \\
        SMT-S~\cite{SMT} & 20 & 4.8 & 356 & 83.7 \\
        BiFormer-S~\cite{biformer} & 26 & 4.5 & 766 & 83.8 \\
        \rowcolor{gray!30}SEC-Swin-T & 29 & 4.8 & 1482 & 83.8 \\
        \rowcolor{gray!30}SECViT-S & 27 & 4.6 & 998 & 84.3 \\
        \midrule[0.5pt]
        Swin-S~\cite{SwinTransformer} & 50 & 8.8 & 1006 & 83.0 \\
        SGFormer-M~\cite{sgformer} & 39 & 7.5 & 598 & 84.1 \\
        SMT-B~\cite{SMT} & 32 & 7.7 & 237 & 84.3 \\
        BiFormer-B~\cite{biformer} & 57 & 9.8 & 498 & 84.3 \\
        MaxViT-S~\cite{maxvit} & 69 & 11.7 & 546 & 84.5 \\
        CMT-B~\cite{cmt} & 46 & 9.3 & 447 & 84.5 \\
        iFormer-B~\cite{iformer} & 48 & 9.4 & 688 & 84.6 \\
        RMT-B~\cite{fan2023rmt} & 54 & 9.7 & 430 & 85.0 \\
        \rowcolor{gray!30}SEC-Swin-S & 50 & 9.2 & 804 & 85.0 \\
        \rowcolor{gray!30}SECViT-B & 57 & 9.8 & 504 & 85.2 \\
        \midrule[0.5pt]
        Swin-B~\cite{SwinTransformer} & 88 & 15.5 & 768 & 83.5 \\
        CSWin-B~\cite{cswin} & 78 & 15.0 & 660 & 84.2 \\
        SMT-L~\cite{SMT} & 80 & 17.7 & 158 & 84.6 \\
        SGFormer-B~\cite{sgformer} & 78 & 15.6 & 388 & 84.7 \\
        iFormer-L~\cite{iformer} & 87 & 14.0 & 410 & 84.8 \\
        MaxViT-B~\cite{maxvit} & 120 & 23.4 & 306 & 84.9 \\
        \rowcolor{gray!30}SEC-Swin-B & 88 & 16.2 & 696 & 85.3 \\
        \rowcolor{gray!30}SECViT-L & 101 & 18.2 & 398 & 85.7 \\
        \bottomrule[1pt]
    \end{tabular}}
    \vspace{-3mm}
    \caption{Comparison of models' efficiency. Throughputs are measured on a single A100 with the batch size of 64.}
    \vspace{-3mm}
    \label{tab:efficiency}
\end{table*}

\paragraph{Different Methods for Merging Vision Tokens.}For MLLM, SEC uses an interleaved merge token approach to reduce the number of vision tokens. Conversely, we also explore a sequential merge token method to achieve a similar reduction. The comparison of these two methods is shown in Tab.~\ref{tab:ab_order_mllm}. The direct sequential merge token approach may result in the loss of critical visual information, significantly degrading the model's performance.

\begin{table*}[t]
    \centering
    \scalebox{1.00}{
    \begin{tabular}{c|ccc|ccccc}
    \toprule[1pt]
    Method & V-T num & Time & Speed & TextVQA & GQA & VQAv2 & POPE & MM-Vet  \\
    \midrule[0.5pt]
    Interleaved & 288+1 & 14h & $1.5\times$ & 60.1 & 63.5 & 78.9 & 87.7 & 33.2 \\
    Sequential & 288+1 & 14h & $1.5\times$ & 52.8(\textcolor{blue}{-7.3}) & 57.1(\textcolor{blue}{-6.2}) & 75.7(\textcolor{blue}{-3.2}) & 81.7(\textcolor{blue}{-6.0}) & 27.6(\textcolor{blue}{-5.6}) \\
    \midrule[0.5pt]
    Interleaved & 144+1 & 10h & $2.1\times$ & 56.8 & 62.0 & 78.0 & 86.1 & 31.7 \\
    Sequential & 144+1 & 10h & $2.1\times$ & 47.2(\textcolor{blue}{-9.6}) & 53.6(\textcolor{blue}{-8.4}) & 71.7(\textcolor{blue}{-6.3}) & 80.0(\textcolor{blue}{-6.1}) & 22.3(\textcolor{blue}{-9.6}) \\
    \bottomrule[1pt]
    \end{tabular}}
    \vspace{-3mm}
    \caption{Different methods for merging vision tokens.}
    \vspace{-3mm}
    \label{tab:ab_order_mllm}
\end{table*}

\section{More Clustering Results for Complex Scenes}
\begin{figure*}
    \centering
    \includegraphics[width=0.99\linewidth]{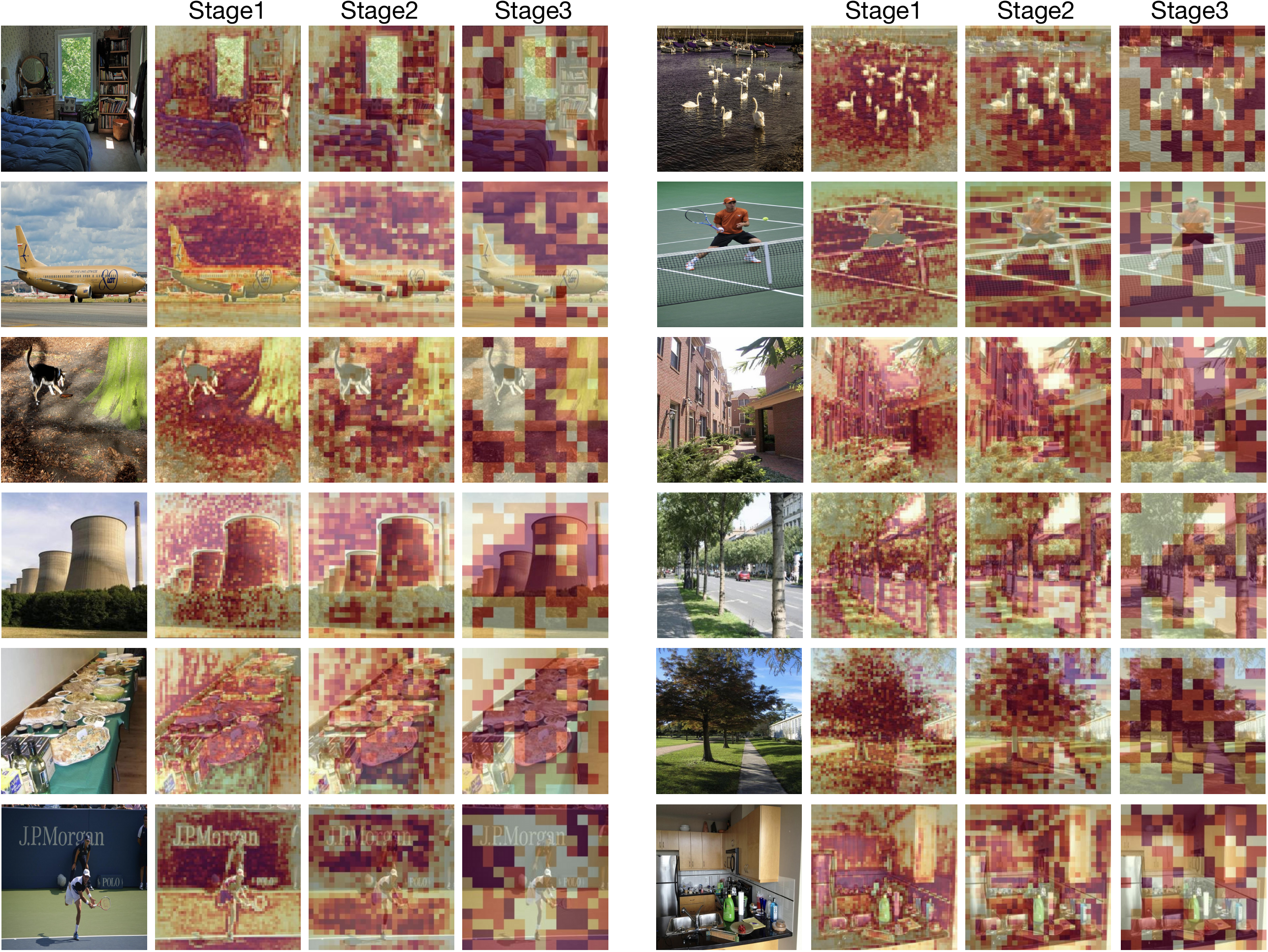}
    \vspace{-3mm}
    \caption{Visualization results for complex scenes.}
    \vspace{-3mm}
    \label{fig:complexscene}
\end{figure*}

To further illustrate the mechanism of SEC, we visualize more images in complex scenes and their clustering results, as shown in Fig.~\ref{fig:complexscene}. Specifically, we visualize the clustering results of the first three stages of SECViT. The results further demonstrate that SEC can better learn fine-grained representations in the shallow layers of the model and semantic representations in the deeper layers.

\end{appendices}

%% file: main.bbl
\begin{thebibliography}{61}
\providecommand{\natexlab}[1]{#1}
\providecommand{\url}[1]{\texttt{#1}}
\expandafter\ifx\csname urlstyle\endcsname\relax
  \providecommand{\doi}[1]{doi: #1}\else
  \providecommand{\doi}{doi: \begingroup \urlstyle{rm}\Url}\fi

\bibitem[Bai et~al.(2023)Bai, Bai, Yang, Wang, Tan, Wang, Lin, Zhou, and Zhou]{bai2023qwenvl}
Jinze Bai, Shuai Bai, Shusheng Yang, Shijie Wang, Sinan Tan, Peng Wang, Junyang Lin, Chang Zhou, and Jingren Zhou.
\newblock Qwen-vl: A versatile vision-language model for understanding, localization, text reading, and beyond, 2023.

\bibitem[Cai and Vasconcelos(2018)]{cai18cascadercnn}
Zhaowei Cai and Nuno Vasconcelos.
\newblock Cascade r-cnn: Delving into high quality object detection.
\newblock In \emph{CVPR}, 2018.

\bibitem[Cha et~al.(2024)Cha, Kang, Mun, and Roh]{honeybee}
Junbum Cha, Wooyoung Kang, Jonghwan Mun, and Byungseok Roh.
\newblock Honeybee: Locality-enhanced projector for multimodal llm.
\newblock In \emph{CVPR}, 2024.

\bibitem[Chen et~al.(2019)Chen, Wang, Pang, et~al.]{mmdetection}
Kai Chen, Jiaqi Wang, Jiangmiao Pang, et~al.
\newblock {MMDetection}: Open mmlab detection toolbox and benchmark.
\newblock \emph{arXiv preprint arXiv:1906.07155}, 2019.

\bibitem[Chen et~al.(2023)Chen, Zhang, Zeng, Zhang, Zhu, and Zhao]{chen2023shikra}
Keqin Chen, Zhao Zhang, Weili Zeng, Richong Zhang, Feng Zhu, and Rui Zhao.
\newblock Shikra: Unleashing multimodal llm's referential dialogue magic, 2023.

\bibitem[Chu et~al.(2023)Chu, Tian, Zhang, Wang, and Shen]{CPVT}
Xiangxiang Chu, Zhi Tian, Bo Zhang, Xinlong Wang, and Chunhua Shen.
\newblock Conditional positional encodings for vision transformers.
\newblock In \emph{ICLR}, 2023.

\bibitem[Contributors(2020)]{mmsegmentation}
MMSegmentation Contributors.
\newblock Mmsegmentation, an open source semantic segmentation toolbox, 2020.

\bibitem[Dai et~al.(2023)Dai, Li, Li, Tiong, Zhao, Wang, Li, Fung, and Hoi]{dai2023instructblip}
Wenliang Dai, Junnan Li, Dongxu Li, Anthony Meng~Huat Tiong, Junqi Zhao, Weisheng Wang, Boyang Li, Pascale Fung, and Steven Hoi.
\newblock Instructblip: Towards general-purpose vision-language models with instruction tuning, 2023.

\bibitem[Ding et~al.(2022)Ding, Xiao, Codella, et~al.]{davit}
Mingyu Ding, Bin Xiao, Noel Codella, et~al.
\newblock Davit: Dual attention vision transformers.
\newblock In \emph{ECCV}, 2022.

\bibitem[Dong et~al.(2022)Dong, Bao, Chen, et~al.]{cswin}
Xiaoyi Dong, Jianmin Bao, Dongdong Chen, et~al.
\newblock Cswin transformer: A general vision transformer backbone with cross-shaped windows.
\newblock In \emph{CVPR}, 2022.

\bibitem[Dosovitskiy et~al.(2021)Dosovitskiy, Beyer, Kolesnikov, et~al.]{vit}
Alexey Dosovitskiy, Lucas Beyer, Alexander Kolesnikov, et~al.
\newblock An image is worth 16x16 words: Transformers for image recognition at scale.
\newblock In \emph{ICLR}, 2021.

\bibitem[Fan et~al.(2023)Fan, Huang, Zhou, and He]{FAT}
Qihang Fan, Huaibo Huang, Xiaoqiang Zhou, and Ran He.
\newblock Lightweight vision transformer with bidirectional interaction.
\newblock In \emph{NeurIPS}, 2023.

\bibitem[Fan et~al.(2024)Fan, Huang, Chen, Liu, and He]{fan2023rmt}
Qihang Fan, Huaibo Huang, Mingrui Chen, Hongmin Liu, and Ran He.
\newblock Rmt: Retentive networks meet vision transformers.
\newblock In \emph{CVPR}, 2024.

\bibitem[Gao et~al.(2023)Gao, Han, Zhang, Lin, Geng, Zhou, Zhang, Lu, He, Yue, Li, and Qiao]{gao2023llamaadapter}
Peng Gao, Jiaming Han, Renrui Zhang, Ziyi Lin, Shijie Geng, Aojun Zhou, Wei Zhang, Pan Lu, Conghui He, Xiangyu Yue, Hongsheng Li, and Yu Qiao.
\newblock Llama-adapter v2: Parameter-efficient visual instruction model, 2023.

\bibitem[guided Transformer~with Evolving Token~Reallocation(2023)]{sgformer}
SG-Former:~Self guided Transformer~with Evolving Token~Reallocation.
\newblock Sucheng ren, xingyi yang, songhua liu, xinchao wang.
\newblock In \emph{ICCV}, 2023.

\bibitem[Guo et~al.(2022{\natexlab{a}})Guo, Han, Wu, Xu, Tang, Xu, and Wang]{cmt}
Jianyuan Guo, Kai Han, Han Wu, Chang Xu, Yehui Tang, Chunjing Xu, and Yunhe Wang.
\newblock Cmt: Convolutional neural networks meet vision transformers.
\newblock In \emph{CVPR}, 2022{\natexlab{a}}.

\bibitem[Guo et~al.(2022{\natexlab{b}})Guo, Lu, Liu, Cheng, and Hu]{VAN}
Meng-Hao Guo, Cheng-Ze Lu, Zheng-Ning Liu, Ming-Ming Cheng, and Shi-Min Hu.
\newblock Visual attention network.
\newblock \emph{arXiv preprint arXiv:2202.09741}, 2022{\natexlab{b}}.

\bibitem[Han et~al.(2024)Han, Wang, Xia, Han, Pu, Ge, Song, Song, Zheng, and Huang]{MLLA}
Dongchen Han, Ziyi Wang, Zhuofan Xia, Yizeng Han, Yifan Pu, Chunjiang Ge, Jun Song, Shiji Song, Bo Zheng, and Gao Huang.
\newblock Demystify mamba in vision: A linear attention perspective.
\newblock In \emph{NeurIPS}, 2024.

\bibitem[Hassani et~al.(2023)Hassani, Walton, Li, Li, and Shi]{NAT}
Ali Hassani, Steven Walton, Jiachen Li, Shen Li, and Humphrey Shi.
\newblock Neighborhood attention transformer.
\newblock In \emph{CVPR}, 2023.

\bibitem[Hatamizadeh et~al.(2023)Hatamizadeh, Yin, Heinrich, Kautz, and Molchanov]{globalvit}
Ali Hatamizadeh, Hongxu Yin, Greg Heinrich, Jan Kautz, and Pavlo Molchanov.
\newblock Global context vision transformers.
\newblock In \emph{ICML}, 2023.

\bibitem[He et~al.(2017)He, Gkioxari, Doll{\'{a}}r, and Girshick]{maskrcnn}
Kaiming He, Georgia Gkioxari, Piotr Doll{\'{a}}r, and Ross~B. Girshick.
\newblock Mask r-cnn.
\newblock In \emph{ICCV}, 2017.

\bibitem[Huang et~al.(2023)Huang, Zhou, Cao, He, and Tan]{stvit}
Huaibo Huang, Xiaoqiang Zhou, Jie Cao, Ran He, and Tieniu Tan.
\newblock Vision transformer with super token sampling.
\newblock In \emph{CVPR}, 2023.

\bibitem[Jaegle et~al.(2021)Jaegle, Gimeno, Brock, Zisserman, Vinyals, and Carreira]{perceiver}
Andrew Jaegle, Felix Gimeno, Andrew Brock, Andrew Zisserman, Oriol Vinyals, and Joao Carreira.
\newblock Perceiver: General perception with iterative attention, 2021.

\bibitem[Jiang et~al.(2021)Jiang, Hou, Yuan, Zhou, Shi, Jin, Wang, and Feng]{tokenlabel}
Zi-Hang Jiang, Qibin Hou, Li Yuan, Daquan Zhou, Yujun Shi, Xiaojie Jin, Anran Wang, and Jiashi Feng.
\newblock All tokens matter: Token labeling for training better vision transformers.
\newblock In \emph{NeurIPS}, 2021.

\bibitem[Kim et~al.(2024)Kim, Seo, Schmid, and Cho]{structvit}
Manjin Kim, Paul~Hongsuck Seo, Cordelia Schmid, and Minsu Cho.
\newblock Learning correlation structures for vision transformers.
\newblock In \emph{CVPR}, 2024.

\bibitem[Kirillov et~al.(2019)Kirillov, Girshick, He, and Doll{\'{a}}r]{semanticfpn}
Alexander Kirillov, Ross Girshick, Kaiming He, and Piotr Doll{\'{a}}r.
\newblock Panoptic feature pyramid networks.
\newblock In \emph{CVPR}, 2019.

\bibitem[Lauren{\c{c}}on et~al.(2024)Lauren{\c{c}}on, Saulnier, Tronchon, Bekman, Singh, Lozhkov, Wang, Karamcheti, Rush, Kiela, et~al.]{laurenccon2024obelics}
Hugo Lauren{\c{c}}on, Lucile Saulnier, L{\'e}o Tronchon, Stas Bekman, Amanpreet Singh, Anton Lozhkov, Thomas Wang, Siddharth Karamcheti, Alexander Rush, Douwe Kiela, et~al.
\newblock Obelics: An open web-scale filtered dataset of interleaved image-text documents.
\newblock In \emph{NeurIPS}, 2024.

\bibitem[Lee et~al.(2022)Lee, Kim, Willette, and Hwang]{mpvit}
Youngwan Lee, Jonghee Kim, Jeffrey Willette, and Sung~Ju Hwang.
\newblock Mpvit: Multi-path vision transformer for dense prediction.
\newblock In \emph{CVPR}, 2022.

\bibitem[Li et~al.(2023)Li, Li, Savarese, and Hoi]{li2023blip}
Junnan Li, Dongxu Li, Silvio Savarese, and Steven Hoi.
\newblock Blip-2: Bootstrapping language-image pre-training with frozen image encoders and large language models.
\newblock In \emph{ICML}, 2023.

\bibitem[Li et~al.(2022)Li, Wang, Gao, Song, Liu, Li, and Qiao]{uniformer}
Kunchang Li, Yali Wang, Peng Gao, Guanglu Song, Yu Liu, Hongsheng Li, and Yu Qiao.
\newblock Uniformer: Unified transformer for efficient spatiotemporal representation learning, 2022.

\bibitem[Li et~al.(2024)Li, Wang, Liu, Tan, Lin, Wu, Chen, Zheng, and Li]{iclr2024MogaNet}
Siyuan Li, Zedong Wang, Zicheng Liu, Cheng Tan, Haitao Lin, Di Wu, Zhiyuan Chen, Jiangbin Zheng, and Stan~Z. Li.
\newblock Moganet: Multi-order gated aggregation network.
\newblock In \emph{ICLR}, 2024.

\bibitem[Liang et~al.(2022)Liang, Ge, Tong, Song, Wang, and Xie]{evit}
Youwei Liang, Chongjian Ge, Zhan Tong, Yibing Song, Jue Wang, and Pengtao Xie.
\newblock Not all patches are what you need: Expediting vision transformers via token reorganizations.
\newblock In \emph{International Conference on Learning Representations}, 2022.

\bibitem[Lin et~al.(2017)Lin, Goyal, Girshick, and andPiotr Doll{\'{a}}r]{retinanet}
Tsung{-}Yi Lin, Priya Goyal, Ross~B. Girshick, and Kaiming~He andPiotr Doll{\'{a}}r.
\newblock Focal loss for dense object detection.
\newblock In \emph{ICCV}, 2017.

\bibitem[Lin et~al.(2023)Lin, Wu, Chen, Huang, and Jin]{SMT}
Weifeng Lin, Ziheng Wu, Jiayu Chen, Jun Huang, and Lianwen Jin.
\newblock Scale-aware modulation meet transformer.
\newblock In \emph{ICCV}, 2023.

\bibitem[Liu et~al.(2023{\natexlab{a}})Liu, Li, Li, and Lee]{liu2023improvedllava}
Haotian Liu, Chunyuan Li, Yuheng Li, and Yong~Jae Lee.
\newblock Improved baselines with visual instruction tuning, 2023{\natexlab{a}}.

\bibitem[Liu et~al.(2023{\natexlab{b}})Liu, Li, Li, and Lee]{llava1.5}
Haotian Liu, Chunyuan Li, Yuheng Li, and Yong~Jae Lee.
\newblock Improved baselines with visual instruction tuning, 2023{\natexlab{b}}.

\bibitem[Liu et~al.(2022{\natexlab{a}})Liu, Wu, Liu, and Guo]{DGT}
Kai Liu, Tianyi Wu, Cong Liu, and Guodong Guo.
\newblock Dynamic group transformer: A general vision transformer backbone with dynamic group attention.
\newblock In \emph{IJCAI}, 2022{\natexlab{a}}.

\bibitem[Liu et~al.(2021)Liu, Lin, Cao, Hu, Wei, Zhang, Lin, and Guo]{SwinTransformer}
Ze Liu, Yutong Lin, Yue Cao, Han Hu, Yixuan Wei, Zheng Zhang, Stephen Lin, and Baining Guo.
\newblock Swin transformer: Hierarchical vision transformer using shifted windows.
\newblock In \emph{ICCV}, 2021.

\bibitem[Liu et~al.(2022{\natexlab{b}})Liu, Mao, Wu, et~al.]{convnext}
Zhuang Liu, Hanzi Mao, Chao-Yuan Wu, et~al.
\newblock A convnet for the 2020s.
\newblock In \emph{CVPR}, 2022{\natexlab{b}}.

\bibitem[Rao et~al.(2021)Rao, Zhao, Liu, Lu, Zhou, and Hsieh]{dynamicvit}
Yongming Rao, Wenliang Zhao, Benlin Liu, Jiwen Lu, Jie Zhou, and Cho-Jui Hsieh.
\newblock Dynamicvit: Efficient vision transformers with dynamic token sparsification.
\newblock In \emph{NeurIPS}, 2021.

\bibitem[Si et~al.(2022)Si, Yu, Zhou, Zhou, Wang, and YAN]{iformer}
Chenyang Si, Weihao Yu, Pan Zhou, Yichen Zhou, Xinchao Wang, and Shuicheng YAN.
\newblock Inception transformer.
\newblock In \emph{NeurIPS}, 2022.

\bibitem[Tang et~al.(2022)Tang, Zhang, Zhu, et~al.]{quadtree}
Shitao Tang, Jiahui Zhang, Siyu Zhu, et~al.
\newblock Quadtree attention for vision transformers.
\newblock In \emph{ICLR}, 2022.

\bibitem[Touvron et~al.(2021)Touvron, Cord, Douze, et~al.]{deit}
Hugo Touvron, Matthieu Cord, Matthijs Douze, et~al.
\newblock Training data-efficient image transformers \& distillation through attention.
\newblock In \emph{ICML}, 2021.

\bibitem[Tu et~al.(2022)Tu, Talebi, Zhang, Yang, Milanfar, Bovik, and Li]{maxvit}
Zhengzhong Tu, Hossein Talebi, Han Zhang, Feng Yang, Peyman Milanfar, Alan Bovik, and Yinxiao Li.
\newblock Maxvit: Multi-axis vision transformer.
\newblock In \emph{ECCV}, 2022.

\bibitem[Wang et~al.(2021{\natexlab{a}})Wang, Xie, Li, Fan, Song, Liang, Lu, Luo, and Shao]{pvt}
Wenhai Wang, Enze Xie, Xiang Li, Deng-Ping Fan, Kaitao Song, Ding Liang, Tong Lu, Ping Luo, and Ling Shao.
\newblock Pyramid vision transformer: A versatile backbone for dense prediction without convolutions.
\newblock In \emph{ICCV}, 2021{\natexlab{a}}.

\bibitem[Wang et~al.(2022{\natexlab{a}})Wang, Xie, Li, Fan, Song, Liang, Lu, Luo, and Shao]{pvtv2}
Wenhai Wang, Enze Xie, Xiang Li, Deng-Ping Fan, Kaitao Song, Ding Liang, Tong Lu, Ping Luo, and Ling Shao.
\newblock Pvtv2: Improved baselines with pyramid vision transformer.
\newblock \emph{Computational Visual Media}, 8\penalty0 (3):\penalty0 1--10, 2022{\natexlab{a}}.

\bibitem[Wang et~al.(2022{\natexlab{b}})Wang, Yao, Chen, Lin, Cai, He, and Liu]{crossformer}
Wenxiao Wang, Lu Yao, Long Chen, Binbin Lin, Deng Cai, Xiaofei He, and Wei Liu.
\newblock Crossformer: A versatile vision transformer hinging on cross-scale attention.
\newblock In \emph{ICLR}, 2022{\natexlab{b}}.

\bibitem[Wang et~al.(2023)Wang, Dai, Chen, Huang, Li, Zhu, Hu, Lu, Lu, Li, et~al.]{internimage}
Wenhai Wang, Jifeng Dai, Zhe Chen, Zhenhang Huang, Zhiqi Li, Xizhou Zhu, Xiaowei Hu, Tong Lu, Lewei Lu, Hongsheng Li, et~al.
\newblock Internimage: Exploring large-scale vision foundation models with deformable convolutions.
\newblock In \emph{CVPR}, 2023.

\bibitem[Wang et~al.(2021{\natexlab{b}})Wang, Huang, Song, Huang, and Huang]{dynamicvit2}
Yulin Wang, Rui Huang, Shiji Song, Zeyi Huang, and Gao Huang.
\newblock Not all images are worth 16x16 words: Dynamic vision transformers with adaptive sequence length.
\newblock In \emph{NeurIPS}, 2021{\natexlab{b}}.

\bibitem[Woo et~al.(2023)Woo, Debnath, Hu, Chen, Liu, Kweon, and Xie]{ConvNeXtV2}
Sanghyun Woo, Shoubhik Debnath, Ronghang Hu, Xinlei Chen, Zhuang Liu, In~So Kweon, and Saining Xie.
\newblock Convnext v2: Co-designing and scaling convnets with masked autoencoders.
\newblock \emph{arXiv preprint arXiv:2301.00808}, 2023.

\bibitem[Xia et~al.(2022)Xia, Pan, Song, Li, and Huang]{dat}
Zhuofan Xia, Xuran Pan, Shiji Song, Li~Erran Li, and Gao Huang.
\newblock Vision transformer with deformable attention.
\newblock In \emph{CVPR}, 2022.

\bibitem[Xiao et~al.(2018)Xiao, Liu, Zhou, Jiang, and Sun]{upernet}
Tete Xiao, Yingcheng Liu, Bolei Zhou, Yuning Jiang, and Jian Sun.
\newblock Unified perceptual parsing for scene understanding.
\newblock In \emph{ECCV}, 2018.

\bibitem[Xie et~al.(2022)Xie, Zhang, Cao, Lin, Bao, Yao, Dai, and Hu]{simmim}
Zhenda Xie, Zheng Zhang, Yue Cao, Yutong Lin, Jianmin Bao, Zhuliang Yao, Qi Dai, and Han Hu.
\newblock Simmim: A simple framework for masked image modeling.
\newblock In \emph{CVPR}, 2022.

\bibitem[Yang et~al.(2023)Yang, Qiao, Yu, et~al.]{MOAT}
Chenglin Yang, Siyuan Qiao, Qihang Yu, et~al.
\newblock Moat: Alternating mobile convolution and attention brings strong vision models.
\newblock In \emph{ICLR}, 2023.

\bibitem[Yang et~al.(2021)Yang, Li, Zhang, Dai, Xiao, Yuan, and Gao]{focal}
Jianwei Yang, Chunyuan Li, Pengchuan Zhang, Xiyang Dai, Bin Xiao, Lu Yuan, and Jianfeng Gao.
\newblock Focal self-attention for local-global interactions in vision transformers.
\newblock In \emph{NeurIPS}, 2021.

\bibitem[Yang et~al.(2022)Yang, Ma, Wu, Tang, Xiao, Zheng, and Li]{ScalableViT}
Rui Yang, Hailong Ma, Jie Wu, Yansong Tang, Xuefeng Xiao, Min Zheng, and Xiu Li.
\newblock Scalablevit: Rethinking the context-oriented generalization of vision transformer.
\newblock In \emph{ECCV}, 2022.

\bibitem[Ye et~al.(2024)Ye, Xu, Xu, Ye, Yan, Zhou, Wang, Hu, Shi, Shi, Li, Xu, Chen, Tian, Qian, Zhang, Huang, and Zhou]{ye2024mplugowl}
Qinghao Ye, Haiyang Xu, Guohai Xu, Jiabo Ye, Ming Yan, Yiyang Zhou, Junyang Wang, Anwen Hu, Pengcheng Shi, Yaya Shi, Chenliang Li, Yuanhong Xu, Hehong Chen, Junfeng Tian, Qi Qian, Ji Zhang, Fei Huang, and Jingren Zhou.
\newblock mplug-owl: Modularization empowers large language models with multimodality, 2024.

\bibitem[Yue et~al.(2021)Yue, Sun, Kuang, Wei, Torr, Zhang, and Lin]{psvit}
Xiaoyu Yue, Shuyang Sun, Zhanghui Kuang, Meng Wei, Philip~HS Torr, Wayne Zhang, and Dahua Lin.
\newblock Vision transformer with progressive sampling.
\newblock In \emph{ICCV}, 2021.

\bibitem[Zeng et~al.(2022)Zeng, Jin, Liu, Qian, Luo, Ouyang, and Wang]{tcformer}
Wang Zeng, Sheng Jin, Wentao Liu, Chen Qian, Ping Luo, Wanli Ouyang, and Xiaogang Wang.
\newblock Not all tokens are equal: Human-centric visual analysis via token clustering transformer.
\newblock In \emph{CVPR}, 2022.

\bibitem[Zhu et~al.(2023{\natexlab{a}})Zhu, Chen, Shen, Li, and Elhoseiny]{zhu2023minigpt4}
Deyao Zhu, Jun Chen, Xiaoqian Shen, Xiang Li, and Mohamed Elhoseiny.
\newblock Minigpt-4: Enhancing vision-language understanding with advanced large language models, 2023{\natexlab{a}}.

\bibitem[Zhu et~al.(2023{\natexlab{b}})Zhu, Wang, Ke, Zhang, and Lau]{biformer}
Lei Zhu, Xinjiang Wang, Zhanghan Ke, Wayne Zhang, and Rynson Lau.
\newblock Biformer: Vision transformer with bi-level routing attention.
\newblock In \emph{CVPR}, 2023{\natexlab{b}}.

\end{thebibliography}
